\pdfoutput=1
\documentclass[final,10pt,twocolumn,letterpaper]{article}

\usepackage{cvpr}              

\usepackage[accsupp]{axessibility}

\usepackage[dvipsnames]{xcolor}

\definecolor{cvprblue}{rgb}{0.21,0.49,0.74}
\usepackage[pagebackref,breaklinks,colorlinks,citecolor=cvprblue]{hyperref}
\usepackage{tcolorbox}
\usepackage{caption}
\usepackage{multirow}
\usepackage{booktabs}
\usepackage{adjustbox}
\usepackage{xcolor}
\usepackage{colortbl}
\usepackage{makecell}
\usepackage{xcolor}
\usepackage{amssymb}
\usepackage{float}
\usepackage[accsupp]{axessibility}  

\definecolor{grey}{rgb}{0.9, 0.9, 0.9}
\definecolor{mgt}{rgb}{0.8, 0.1, 0.8}

\definecolor{blu}{rgb}{0.1, 0.8, 0.1}

\usepackage{soul}

\definecolor{supervisedBase}{rgb}{0.1, 0.1, 0.8} 
\definecolor{unsupervisedBase}{rgb}{0.8, 0.1, 0.1}
\definecolor{oursColorBase}{rgb}{0.9,0.9,0.9}

\colorlet{supervisedColor}{supervisedBase!10!white}
\colorlet{unsupervisedColor}{unsupervisedBase!10!white}
\colorlet{oursColor}{oursColorBase!50!white}

\definecolor{azure(colorwheel)}{rgb}{0.0, 0.5, 1.0}

\definecolor{codegreen}{rgb}{0,0.6,0}
\definecolor{codegray}{rgb}{0.5,0.5,0.5}
\definecolor{codepurple}{rgb}{0.58,0,0.82}
\definecolor{backcolour}{rgb}{0.95,0.95,0.92}

\usepackage{algorithm}
\usepackage{algorithmic}
\usepackage{listings}

\lstdefinestyle{mystyle}{
    commentstyle=\color{codegreen},
    keywordstyle=\color{magenta},
    numberstyle=\tiny\color{codegray},
    stringstyle=\color{codepurple},
    basicstyle=\ttfamily\footnotesize,
    breakatwhitespace=false,         
    breaklines=true,                 
    captionpos=b,                    
    keepspaces=true,                 
    numbers=left,                    
    numbersep=5pt,                  
    showspaces=false,                
    showstringspaces=false,
    showtabs=false,                  
    tabsize=2
}

\usepackage{lipsum}
\usepackage{footnote}
\usepackage{subcaption}
\usepackage{ulem}

\makeatletter
\def\blfootnote{\gdef\@thefnmark{}\@footnotetext}
\makeatother
\def\paperID{16506} 
\def\confName{CVPR}
\def\confYear{2025}

\title{GOAL: Global-local Object Alignment Learning}

\author{Hyungyu Choi\textsuperscript{1}\thanks{Authors contributed equally. \hspace{1em} \textsuperscript{$\dagger$}Corresponding author.} , Young Kyun Jang\textsuperscript{2}\footnotemark[1] , Chanho Eom\textsuperscript{1}\textsuperscript{$\dagger$} \\
\\ \text{Chung-Ang University\textsuperscript{1} \quad Meta AI\textsuperscript{2}}\\ \\
\href{https://perceptualai-lab.github.io/GOAL/}{\textit{https://perceptualai-lab.github.io/GOAL}}
}

\begin{document}
\maketitle
\begin{abstract}
Vision-language models like CLIP have shown impressive capabilities in aligning images and text, but they often struggle with lengthy and detailed text descriptions because of their training focus on short and concise captions. We present GOAL (Global-local Object Alignment Learning), a novel fine-tuning method that enhances CLIP's ability to handle lengthy text by leveraging both global and local semantic alignments between image and lengthy text. Our approach consists of two key components: Local Image-Sentence Matching (LISM), which identifies corresponding pairs between image segments and descriptive sentences, and Token Similarity-based Learning (TSL), which efficiently propagates local element attention through these matched pairs. Evaluating GOAL on three new benchmarks for image-lengthy text retrieval, we demonstrate significant improvements over baseline CLIP fine-tuning, establishing a simple yet effective approach for adapting CLIP to detailed textual descriptions. Through extensive experiments, we show that our method's focus on local semantic alignment alongside global context leads to more nuanced and representative embeddings, particularly beneficial for tasks requiring fine-grained understanding of lengthy text descriptions.

\end{abstract}    
\section{Introduction}
\label{sec:intro}

\begin{figure}[!t]
\centering
\subcaptionbox{CLIP
\label{fig:intro1_a}}{\includegraphics[height=6.2cm]{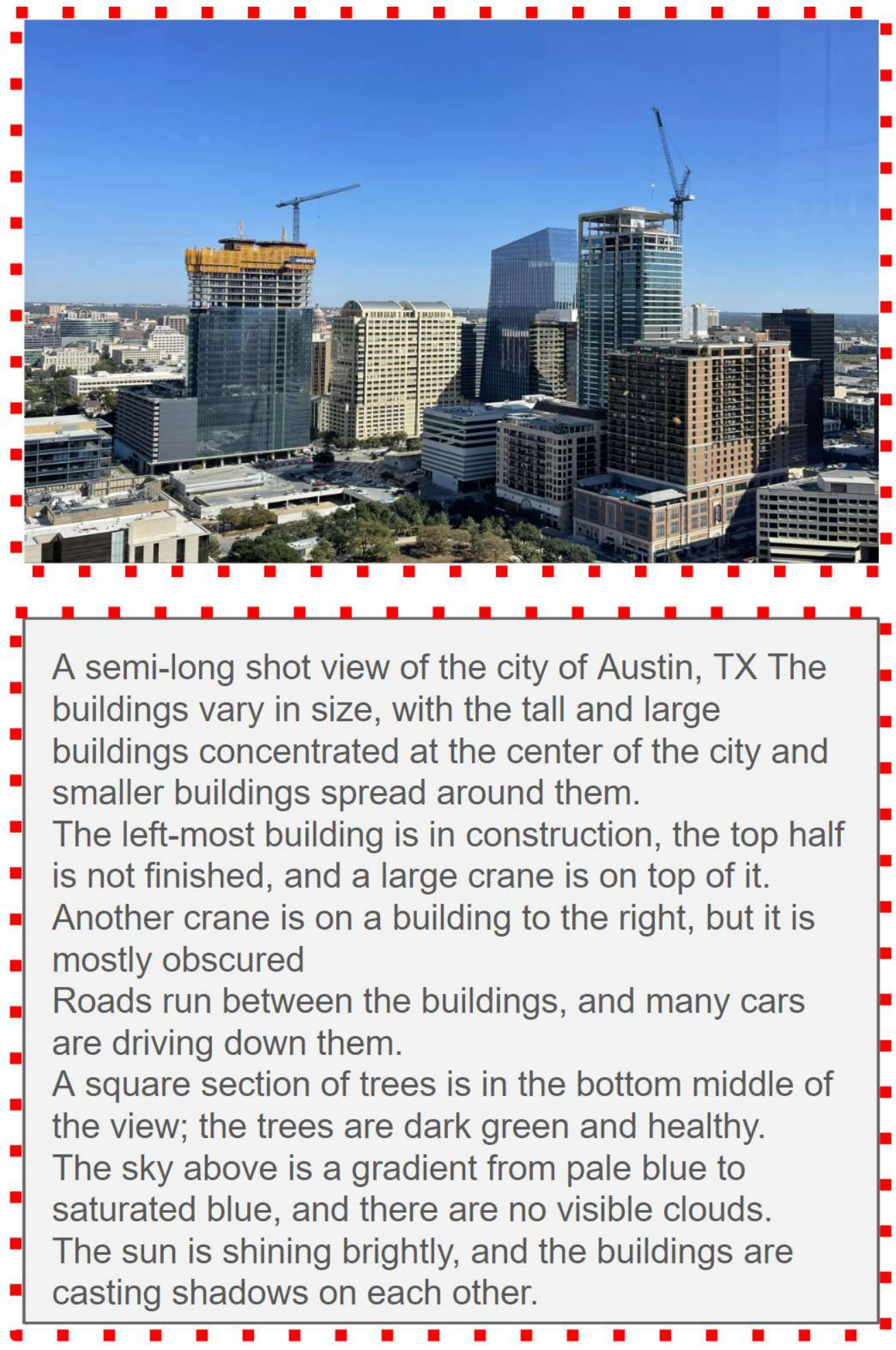}}
\subcaptionbox{GOAL
\label{fig:intro_b}}{\includegraphics[height=6.2cm]{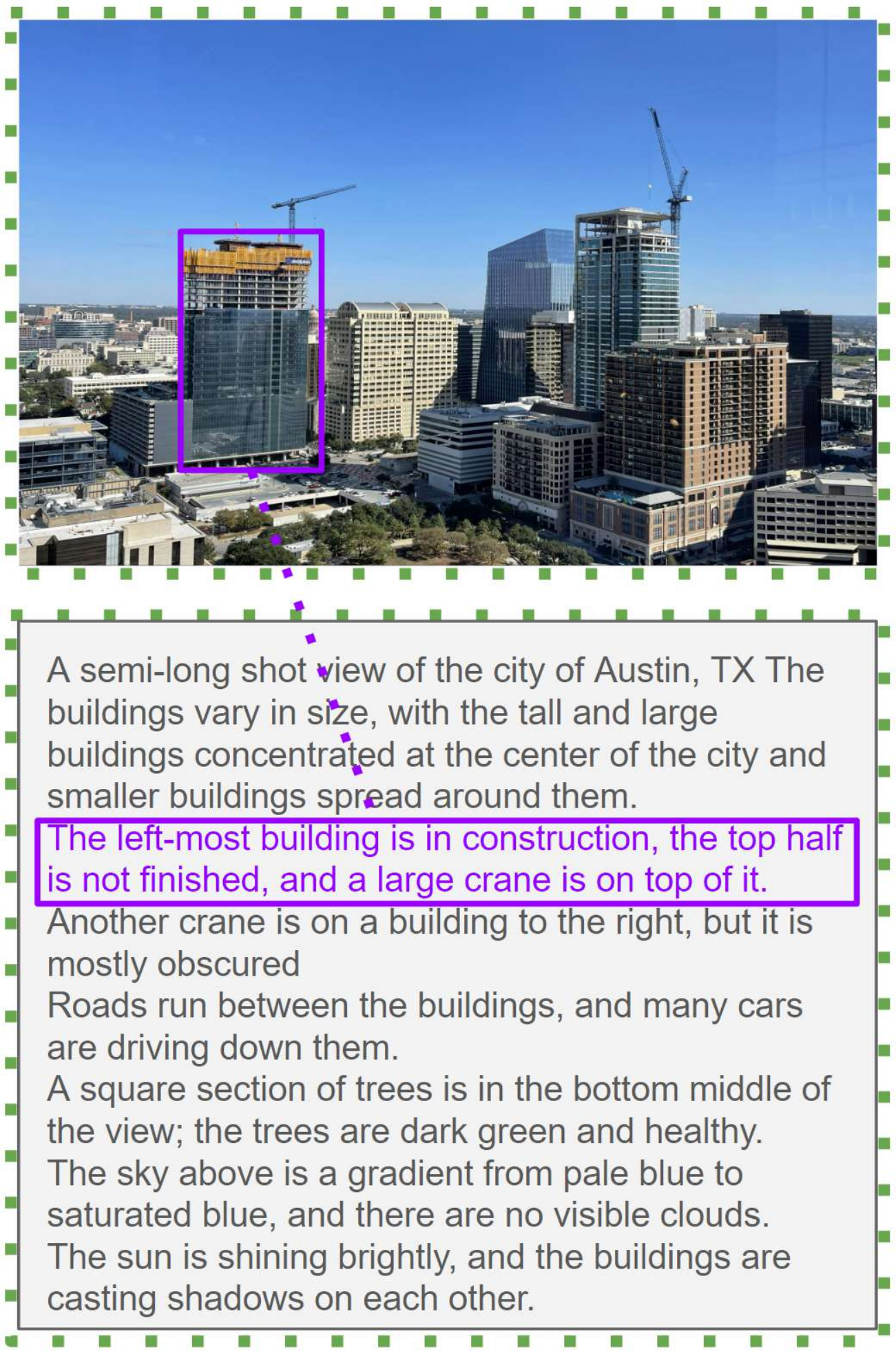}}
\caption{Comparison of CLIP and our GOAL's capability in handling image-text alignment. (a) CLIP is limited to global image-text matching, treating the entire image and full caption as single units without detailed associations. (b) GOAL can establish precise local alignments between specific regions in the image and their corresponding textual descriptions in the caption (highlighted in purple). }
\label{fig:intro}
\end{figure}

After the emergence of CLIP~\cite{CLIP}, numerous methods ~\cite{sclip}\cite{siglip}\cite{maskclip}\cite{llip} have been proposed to bridge the modality gap between images and text showcasing significant advancements. By aligning hundreds of millions of image-caption pairs through contrastive learning, CLIP successfully encodes images and text into a unified embedding space. The resulting distribution of image and text embeddings captures both visual and linguistic semantics, enabling zero-shot transfer to various downstream tasks, such as classification~\cite{vgg}\cite{resnet}\cite{inception}\cite{resnext} and retrieval~\cite{vis-w2v}\cite{odyssey}\cite{ren}, while achieving decent performance.

However, fine-tuning a pre-trained CLIP (Fig.~\ref{fig:intro} (a)) model for specific domains faces limitations, as CLIP is trained on general, short captions (\textit{e.g.}, 77 tokens in the vanilla model) that focus on high-level image concepts. When tasked with longer, more detailed text, CLIP struggles to capture nuanced information, as the unified embedding space is optimized for concise descriptions. This makes adapting CLIP for retrieval tasks requiring lengthy text challenging without architectural adjustments or specialized training techniques.

In this paper, we propose a novel but simple fine-tuning method for image and lengthy text pairs, called \textbf{G}lobal-local \textbf{O}bject \textbf{A}lignment \textbf{L}earning (GOAL) (Fig.~\ref{fig:intro} (b)). Here, we refer to \textit{``global"} as the entire image or text and \textit{``local"} as a sub-part, such as a segment of the image or a specific sentence in the text. The idea behind GOAL is to enable the encoder model to focus on the dominant local elements within each image and text sample, thereby enhancing the overall understanding of the sample and producing a more representative embedding. 

GOAL has two key components: First, \textbf{L}ocal \textbf{I}mage-\textbf{S}entence \textbf{M}atching (LISM), a pipeline that extracts local segments from images and matches them with corresponding descriptive sentences from the entire caption. Second, we introduce \textbf{T}oken \textbf{S}imilarity-based \textbf{L}earning (TSL), a method that effectively propagates attention of local element using the local pairs obtained through the LISM pipeline. To address the challenge of image-lengthy text retrieval, we propose new benchmarks, evaluating GOAL on three diverse datasets (DOCCI~\cite{DOCCI}, DCI~\cite{DCI}, and Urban1k~\cite{LongCLIP} ) containing image-lengthy caption pairs, and demonstrating substantial fine-tuning performance compared to the original CLIP tuning.
 The main contributions of our work can be summarized as follows:
 
\begin{itemize}
\item We propose GOAL, a fine-tuning approach that enhances CLIP's understanding of local elements within samples to improve embedding representations.

\item GOAL includes two components: Local Image-Sentence Matching (LISM) for generating pseudo local pairs, and Token Similarity-based Learning (TSL) for efficient propagation the attention of local elements.

\item Through experiments on newly proposed benchmarks, we show that GOAL significantly improves performance over the original CLIP and baseline models.
\end{itemize}
\section{Related Work}
\label{sec:Related Work}

\paragraph{Vision-Language Pre-training.}
Research on addressing alignment differences between vision and language modalities has brought the Contrastive Language-Image Pre-training (CLIP)~\cite{CLIP} model into the spotlight. CLIP, a multi-modal embedding model trained through contrastive learning on over 400 million image-text pairs, effectively aligns visual and textual representations while demonstrating remarkable zero-shot capabilities. Following its success, larger pre-training models emerged, such as ALIGN~\cite{ALIGN} and Florence~\cite{Florence}, trained on image-text pairs from datasets containing 1.8B and 900M samples, respectively. However, these models typically rely on short, broad image descriptions as captions, causing them to miss crucial local-level detailed information. While Long-CLIP~\cite{LongCLIP} addressed this limitation by utilizing synthetic lengthy captions generated by multi-modal LLMs~\cite{survey}\cite{freeze}\cite{vita}\cite{Video-mme}, it requires an expensive data preparation process. To overcome this limitation more efficiently, we present a fine-tuning method that enhances CLIP's ability to capture both local-detail and global-semantic information by training it on a dataset containing detailed, multi-sentence captions.

\paragraph{Utilizing Local Elements in Vision-Language Model Training.} 
In terms of vision-language alignment models, using local elements' knowledge to improve the model's general ability has been widely explored across various domains. Visual-Textual Attributes Alignment (ViTAA)~\cite{ViTAA} learns to align full-person images corresponding to the global-level with text describing the whole person to perform a person re-identification task~\cite{reid1}\cite{reid2}\cite{reid3}\cite{reid4}, while also learning to align the image and text for attributes (\textit{e.g.}, hair, pants, shoes) that correspond to the local-level. This approach combines global-local relations, enabling richer visual-language representation learning. CLOC (Contrastive Localized Language-Image Pre-Training)~\cite{CLOC} builds 2 billion image-text datasets and uses them for pre-training models by matching local objects and phrase-levels through Open-vocabulary Detector (\textit{e.g.}, OWLv2~\cite{OWLv2}, GLIPv2~\cite{GLIPv2}) models to improve localization capabilities while maintaining CLIP's global-level representation, demonstrating superior performance compared to the original pre-trained CLIP model. In contrast, our proposed GOAL method efficiently learns global-local relationships through fine-tuning with significantly fewer datasets and computational resources compared to large-scale pre-training approaches.
\section{Method}
\label{sec:Method}

In this section, we introduce Local Image-Sentence Matching (LISM), a pipline that  generates local-level pseudo pairs from a given image-caption pair~(Sec.~\ref{subsec:Local Image-Sentence Matching}). We then  present the Token Similarity-based Learning (TSL) method, which leverages these pseudo pairs to address global-level biases in CLIP~\cite{CLIP}~(Sec.~\ref{subsec:Token Similarity based Learning}).

\subsection{Local Image Sentence Matching}
\label{subsec:Local Image-Sentence Matching}
\begin{figure*}[!t]
\centering
\includegraphics[width=\textwidth]{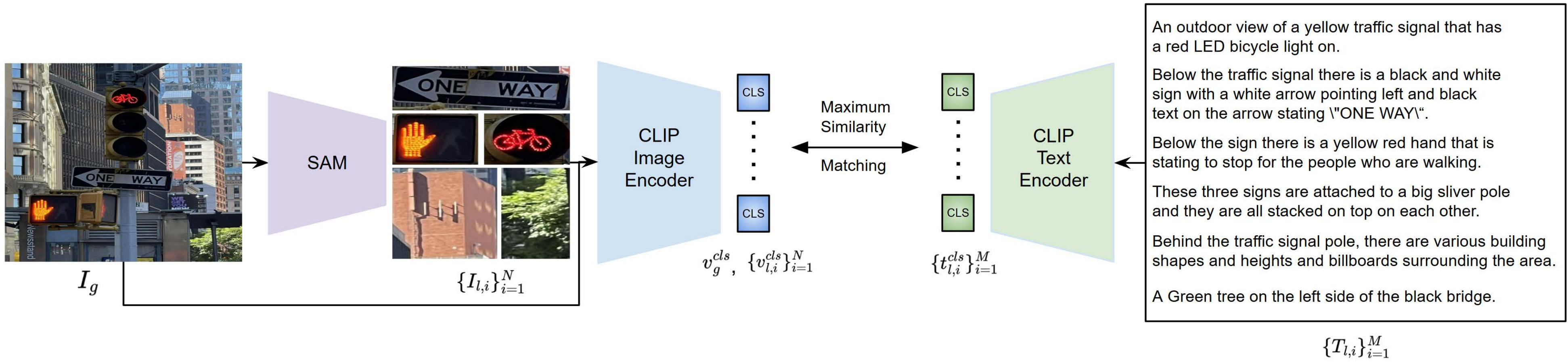}
\caption{Overview of Local Image-Sentence Matching (LISM) pipeline. Given a global image and its detailed caption, LISM uses SAM to segment the image into local regions and splits the caption into individual sentences. These local pairs are then processed through CLIP encoders to obtain CLS embeddings, which are used for maximum similarity matching to identify the most relevant image-sentence pairs. }
\label{fig:LISM}
\end{figure*}

We propose Local Image-Sentence Matching (LISM) Fig.~\ref{fig:LISM}, which separates a given caption into individual sentences and identifies corresponding image segments, matching each sentence with its relevant segment. To this end, we first decompose a given caption $T_g$, which provides detailed descriptions of a given image $I_g$, into individual sentences, resulting in text segments $\{T_{l,i}\}_{i=1}^M$, where $M$ is the number of sentences. We then leverage SAM~\cite{SAM} to segment the image $I_g$ into semantic units, obtaining masks for individual objects along with the background. We expand each mask into a rectangular bounding box that includes the surrounding area, allowing us to leverage contextual information for matching with the caption. As a result, we obtain a set of local images, $\{I_{l,i}\}_{i=1}^N$, where $N$ represents the number of local regions. Note that in this process, we filter out segments smaller than 1\% of the total image area to exclude very small objects and reduce noise from SAM. 

We use CLIP~\cite{CLIP} to match the decomposed caption segments with the corresponding image segments. Specifically, we extract the CLS token embeddings for each local text segment $T_{l,j}$ from the text encoder of CLIP, $\phi_t$ as follows: 
\begin{equation}
    \{t^{cls}_{l,i}\}_{i=1}^M = \phi_t(\{T_{l,i}\}_{i=1}^M).
\end{equation}
Similarly, for both the original image $I_g$ and each image segment $I_{l,i}$, we extract the CLS token embeddings from the visual encoder of CLIP as follows: 
\begin{equation}
\begin{aligned}
    v^{cls}_g = \phi_v(I_g), \quad \{v^{cls}_{l,i}\}_{i=1}^N &= \phi_v(I_{l,i}). \quad
\end{aligned}
\end{equation}
Next, we compute the cosine similarity between each local text embedding $t^{cls}_{l,i}$ and the global image embedding $v^{cls}_g$ or the local image embeddings $\{{v^{cls}_{l,i}}\}_{i=1}^N$. Among all matched pairs, each local text embedding is matched with its highest similarity image embedding. From all these matched pairs, we select the one pair with the highest similarity score and denote it as $(I_l, T_l)$. If the matched image in this selected pair is the global image $I_g$, we discard this pair.  This matching strategy excludes global image matches from the final selection to ensure high-quality local pair associations.

\subsection{Token Similarity based Learning}
\label{subsec:Token Similarity based Learning}
\begin{figure*}[!t]
\centering
\includegraphics[width=\textwidth]{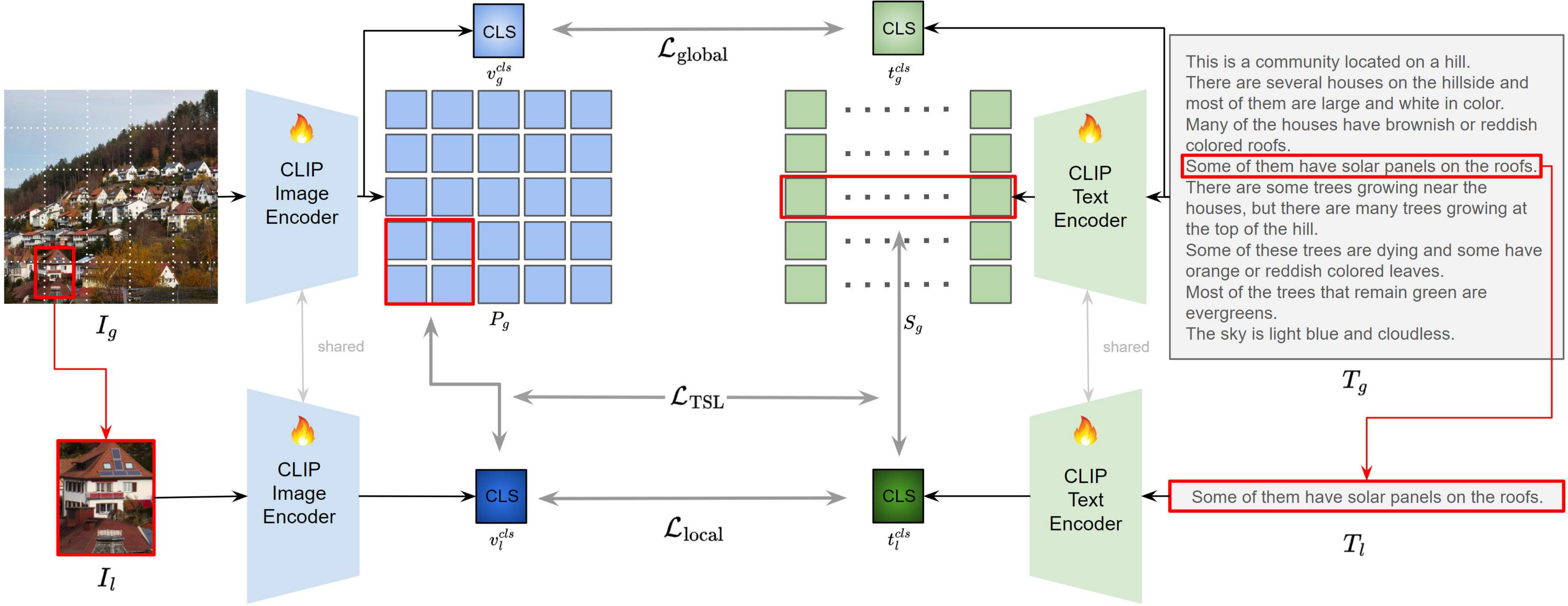}
\caption{Overview of Token Similarity based Learning (TSL). The framework processes global image-text pairs and their local pairs through shared CLIP encoders, extracting patch and sequence tokens. TSL identifies and projects corresponding token regions to match local CLS embeddings, enabling attention on local element.}
\label{fig:TSL}
\end{figure*}

While CLIP's pretraining with image-text pairs effectively learns global alignment, its training with brief captions limits the model's ability to capture fine-grained local details from lengthy descriptions. To address this, we propose Token Similarity based Learning (TSL) (Fig.~\ref{fig:TSL}). Our approach uses local pairs obtained through the LISM pipeline and implements a fine-tuning strategy that effectively propagates local-level information. Specifically, TSL maximizes the similarity between patch tokens of local regions in the global image and their corresponding local image embeddings, while applying the same principle to text by increasing the similarity between sequence tokens of local parts in the global text and their corresponding local text embeddings.
To implement this strategy, we need to extract both local and global features from the input pairs.  Using CLIP's vision encoder $\phi_v$ and text encoder $\phi_t$, we extract both local and global features as follows:
For the local text $T_l$:
\begin{equation}
t^{cls}_l = \phi_t(T_l) \in \mathbb{R}^d,
\end{equation}
where $t^{cls}_l$ represents the last layer CLS token embedding. For the global text $T_g$, the text encoder extracts:
\begin{equation}
S_g = \phi_t(T_g) \in \mathbb{R}^{M \times d},
\end{equation}
where $M$ is the sequence length of $T_g$, and $S_g$ represents the last layer sequence tokens of $T_g$. To handle text sequences longer than CLIP's standard 77 token limit, we adopt Long-CLIP's~\cite{LongCLIP} positional embedding interpolation method in our text encoder.
For the local image $I_l$, we obtain:
\begin{equation}
v^{cls}_l = \phi_v(I_l) \in \mathbb{R}^d,
\end{equation}
where $v^{cls}_l$ represents the last layer CLS token embedding. For the global image $I_g$, the vision encoder extracts:
\begin{equation}
P_g = \phi_v(I_g) \in \mathbb{R}^{N \times d},
\end{equation}
where $N$ denotes the number of patch tokens in $I_g$ , $d$ is the embedding dimension and $P_g$ represents the last layer patch tokens of $I_g$. We process both global and local pairs through shared CLIP encoders to learn both types of features simultaneously. This weight sharing ensures consistent encoding in the shared embedding space. Let $\mathcal{T}$ denote the set of token indices corresponding to the local text segment. We can identify the sequence tokens in $S_g$ that correspond to $T_l$:
\begin{equation}
S_m = \frac{1}{|\mathcal{T}|}\sum_{i \in \mathcal{T}} S_g[i] \in \mathbb{R}^d,
\end{equation}
where $|\mathcal{T}|$ denotes the number of selected sequence tokens. The aggregated features are then projected into a shared embedding space, where both text and image representations are aligned: 
\begin{equation}
\hat{S_l} = \text{\textit{proj}}(S_m) \in \mathbb{R}^d,
\end{equation}
where $\text{\textit{proj}}(\cdot)$ represents a learned projection function.
 
Given that each local image region $I_l$ has its bounding box coordinates $(x_1, y_1, x_2, y_2)$ obtained from LISM in the global image $I_g$, we can leverage this spatial information to identify specific patch tokens from $P_g$ that correspond to the local image region, filtering out patches from other parts of the global image. Let $\mathcal{B}$ denote the set of indices of patch tokens located inside the bounding box. We aggregate these tokens using average pooling to capture comprehensive information from the selected region:
\begin{equation}
P_m = \frac{1}{|\mathcal{B}|}\sum_{i \in \mathcal{B}} P_g[i] \in \mathbb{R}^d,
\end{equation}
where $|\mathcal{B}|$ denotes the number of selected patch tokens. The aggregated features are then projected into a shared embedding space where both text and image representations are aligned: 
\begin{equation}
\hat{P_l} = \text{\textit{proj}}(P_m ) \in \mathbb{R}^d,
\end{equation}
where $\text{\textit{proj}}(\cdot)$ represents a learned projection function.
We train our model with multiple objectives combined into a final loss function:
\begin{equation}
\mathcal{L}_{\text{total}} =  \lambda_{global}\mathcal{L}_{\text{global}} + \lambda_{local}\mathcal{L}_{\text{local}} + \lambda_{TSL}\mathcal{L}_{\text{TSL}},
\end{equation}
where $\lambda$ is a hyperparameter controlling the contribution of local alignment.  We apply contrastive learning at both global and local levels, adopting the contrastive learning used in CLIP . At the global level:
\begin{equation}
\mathcal{L}_{\text{global}} = \mathcal{L}_{\text{contrast}}(v_{g}^{cls}, t_{g}^{cls}),
\end{equation}
where  $v_{g}^{cls}$ and  $t_{g}^{cls}$ are the CLS token embeddings of the global image $I_g$ and global text $T_g$, respectively. This global alignment ensures that the model maintains CLIP's original capability to capture global relationships between image-text pairs. Similarly, for local-level contrastive learning:
\begin{equation}
\mathcal{L}_{\text{local}} = \mathcal{L}_{\text{contrast}}(v_{l}^{cls},
t_{{l}}^{cls}),
\end{equation}
where  $v_{l}^{cls}$ and  $t_{l}^{cls}$ are the CLS token embeddings of the local image $I_l$ and local text $T_l$, respectively. By applying contrastive learning to local CLS token pairs, we encourage precise alignment between local image regions and their corresponding textual descriptions, enabling the model to learn cross-modal relationships.

The token similarity loss $\mathcal{L}_{\text{TSL}}$ maximizes the similarity between projected tokens and their corresponding local CLS token embeddings for both image and text:
\begin{equation}
\mathcal{L}_{\text{TSL}} = \text{MSE}(\text{\textit{sim}}(\hat{P_l},v_{l}^{cls}), \mathbf{1}) + \text{MSE}(\text{\textit{sim}}(\hat{S_l},t_{{l}}^{cls}), \mathbf{1}),
\end{equation}
where $\text{\textit{sim}}(\cdot)$ denotes a function that computes an $n \times n$ similarity matrix with $n$ being the batch size, and $\mathbf{1}$ is a $n \times n$ matrix with ones on its diagonal entries. By optimizing this loss, the model learns to maximize the similarity between local CLS token embeddings and their corresponding regions in global tokens. This token-level alignment strategy enables the model to attention on local element, enhancing fine-grained understanding capabilities.
This fine-tuning method effectively addresses CLIP's inherent limitation in capturing local details from lengthy descriptions, which stems from its pre-training with brief captions. Through the combination of token-level similarity learning and global-local contrastive learning, our approach enables comprehensive understanding of cross-modal relationships with attention on local element from detailed text descriptions.

\section{Experiments}

In this section, we present our experimental setup in Sec.~\ref{sec:Experimental setup}. Our ablation study in Sec.~\ref{sec:ablation} demonstrates the effectiveness of each component in our framework through experiments. We provide zero-shot experimental results in Sec.~\ref{sec:comparison} to show our model's generalization capability across different datasets. Finally, we present qualitative analysis in Sec.~\ref{sec:Qualitative results} through visualization of attention maps.

\subsection{Experimental setup}
\label{sec:Experimental setup}

\paragraph{Dataset.} 
We conduct experiments on three datasets: DOCCI~\cite{DOCCI}, DCI~\cite{DCI} and Urban1k~\cite{LongCLIP}, each containing images with long and detailed captions, designed to enable vision-language models to learn fine-grained visual-textual relationships. The DOCCI dataset consists of 9,647 training samples and a combined test set of 5,100 samples (5,000 from the test set and 100 from the qualification-test set). Since DCI's original test set contains only 100 samples, we instead sampled 2,000 examples from its training set of 7,805 samples to create a larger test set, establishing a train-test ratio similar to DOCCI. For both datasets, we generate pseudo local pairs through our LISM. The datasets and our sampled test sets used in this research are publicly available on GitHub\footnote{\href{https://github.com/PerceptualAI-Lab/GOAL/tree/main/datasets}{https://github.com/PerceptualAI-Lab/GOAL/tree/main/datasets}}.

\paragraph{Training setting.} 
To validate our approach, we conduct experiments using two different CLIP~\cite{CLIP} backbone architectures: ViT-B/16, and ViT-L/14~\cite{attention}\cite{ViT}. Both models are fine-tuned for 10 epochs with a batch size of 16. We set the balance hyperparameters in the total loss function as $\lambda_{global}=1$, $\lambda_{TSL}=1$, and $\lambda_{local}=0.5$ to maintain strong global and TSL learning while moderating the contribution of local loss. The training was performed on a single NVIDIA RTX 4090 GPU for base models and an NVIDIA A6000 GPU for the ViT-L/14 model, taking approximately 1 and 2 hours respectively.

\paragraph{Test setting.} 
To handle the long text sequences during inference, we adopt the positional embedding interpolation technique from Long-CLIP~\cite{LongCLIP}. We evaluate our method on two different test scenarios: the original test set and our proposed global-local test set. For the original test set, we evaluate Text-to-Image (T2I) and Image-to-Text (I2T) retrieval performance using Recall@k. For the second scenario, we create a pseudo global-local test set by applying our proposed LISM to the original test set. Specifically, we generate local pairs for each image-text pair in the original test set and append the local pair with the highest similarity score to create the pseudo global-local test set. For this extended test set, we using mAP@k as our evaluation metric since we need to evaluate retrieval performance in situations with multiple correct answers in our global-local matching scenario. Both global and local texts are considered correct answers when querying with either global or local images, and similarly, both global and local images are considered correct answers when querying with either type of text.

\subsection{Ablation study}
\label{sec:ablation}

\renewcommand{\arraystretch}{0.8}
\begin{table*}[!t]
\centering
\small
\begin{tabular}{c|c|ccc|cccc|cccc}
\toprule
\multirow{2}{*}{Backbone} & \multirow{2}{*}{Methods} & \multicolumn{3}{c|}{Loss} & \multicolumn{4}{c|}{Text to Image Recall@K} & \multicolumn{4}{c}{Image to Text Recall@K} \\
\noalign{\vskip 1pt} 
\cline{3-13}
\noalign{\vskip 2pt} 
& & Global & Local & TSL & R@1 & R@5 & R@25 & R@50 & R@1 & R@5 & R@25 & R@50 \\
\midrule
\multirow{4}{*}{ViT-B/16} 
& Global fine-tuning & \textcolor{black}{\checkmark} & & & \underline{72.41} & 93.27 & \underline{99.31} & 99.76 & \underline{72.04} & 93.37 & \underline{99.35} & \underline{99.80} \\
& Local fine-tuning & & \textcolor{black}{\checkmark} & & 65.82 & 89.96 & 98.37 & 99.39 & 65.73 & 90.35 & 98.35 & 99.51 \\
& w/o TSL & \textcolor{black}{\checkmark} & \textcolor{black}{\checkmark} & & 72.08 & \underline{93.73} & 99.24 & \underline{99.82} & 71.80 & \underline{93.57} & 99.29 & 99.76 \\
& GOAL & \textcolor{black}{\checkmark} & \textcolor{black}{\checkmark} & \textcolor{black}{\checkmark} & \textbf{79.47} & \textbf{96.65} & \textbf{99.69} & \textbf{99.92} & \textbf{79.43} & \textbf{96.14} & \textbf{99.61} & \textbf{99.90} \\
\midrule
\multirow{4}{*}{ViT-L/14} 
& Global fine-tuning & \textcolor{black}{\checkmark} & & & 74.00 & 93.84 & 99.04 & 99.67 & 73.55 & 93.94 & 99.16 & \underline{99.78} \\
& Local fine-tuning & & \textcolor{black}{\checkmark} & & 67.39 & 90.67 & 98.16 & 99.20 & 66.33 & 90.41 & 98.10 & 99.43 \\
& w/o TSL & \textcolor{black}{\checkmark} & \textcolor{black}{\checkmark} & & \underline{74.75} & \underline{94.31} & \underline{99.12} & \underline{99.71} & \underline{74.55} & \underline{94.37} & \underline{99.27} & \underline{99.78} \\
& GOAL & \textcolor{black}{\checkmark} & \textcolor{black}{\checkmark} & \textcolor{black}{\checkmark} & \textbf{84.37} & \textbf{97.55} & \textbf{99.76} & \textbf{99.98} & \textbf{82.57} & \textbf{97.37} & \textbf{99.82} & \textbf{99.98} \\
\bottomrule

\end{tabular}
\vspace{-2mm}
\captionsetup{width=\linewidth}
\caption{Original test set results on DOCCI dataset. Comparison of retrieval performance across different fine-tuning approaches using ViT-B/16 and ViT-L/14 models. The evaluation metrics include both text-to-image and image-to-text Recall@K. The best and second-best scores for each method are marked in \textbf{bold} and \underline{underlined}, respectively.}
\vspace{-0.5mm}
\label{table1}
\end{table*}

\begin{table*}[!t]
\centering
\small
\renewcommand{\arraystretch}{0.8}
\begin{tabular}{c|c|ccc|cccc|cccc}
\toprule
\multirow{2}{*}{Backbone} & \multirow{2}{*}{Methods} & \multicolumn{3}{c|}{Loss} & \multicolumn{4}{c|}{Text to Image Recall@K} & \multicolumn{4}{c}{Image to Text Recall@K} \\
\noalign{\vskip 1pt} 
\cline{3-13}
\noalign{\vskip 2pt} 
& & Global & Local & TSL & R@1 & R@5 & R@25 & R@50 & R@1 & R@5 & R@25 & R@50 \\
\midrule
\multirow{4}{*}{ViT-B/16} 
& Global fine-tuning & \textcolor{black}{\checkmark} & & & 66.43 & \underline{84.74} & \underline{93.80} & \underline{96.10} & \underline{66.58} & 84.74 & \underline{95.10} & 97.65 \\
& Local fine-tuning & & \textcolor{black}{\checkmark} & & 59.38 & 78.49 & 90.70 & 93.85 & 58.18 & 78.74 & 90.05 & 93.75 \\
& w/o TSL & \textcolor{black}{\checkmark} & \textcolor{black}{\checkmark} & & \underline{66.63} & 84.04 & 93.75 & 96.05 & 66.43 & \underline{85.29} & 95.00 & \underline{97.75} \\
& GOAL & \textcolor{black}{\checkmark} & \textcolor{black}{\checkmark} & \textcolor{black}{\checkmark} & \textbf{72.64} & \textbf{89.89} & \textbf{95.95} & \textbf{97.25} & \textbf{72.84} & \textbf{90.50} & \textbf{96.60} & \textbf{97.90} \\
\midrule
\multirow{4}{*}{ViT-L/14} 
& Global fine-tuning & \textcolor{black}{\checkmark} & & & 65.73 & 84.24 & 93.25 & \underline{96.30} & 65.73 & \underline{86.04} & 94.65 & \underline{96.25} \\
& Local fine-tuning & & \textcolor{black}{\checkmark} & & 53.88 & 75.54 & 87.84 & 91.75 & 51.63 & 72.64 & 87.49 & 91.10 \\
& w/o TSL & \textcolor{black}{\checkmark} & \textcolor{black}{\checkmark} & & \underline{66.38} & \underline{84.44} & \underline{93.40} & \underline{96.30} & \underline{66.23} & \underline{86.04} & \underline{94.75} & 96.50 \\
& GOAL & \textcolor{black}{\checkmark} & \textcolor{black}{\checkmark} & \textcolor{black}{\checkmark} & \textbf{76.89} & \textbf{91.05} & \textbf{96.55} & \textbf{97.75} & \textbf{76.59} & \textbf{91.20} & \textbf{96.55} & \textbf{98.25} \\
\bottomrule
\end{tabular}
\vspace{-2mm}
\captionsetup{width=\linewidth}
\caption{Original test set results on DCI dataset. Comparison of retrieval performance across different fine-tuning approaches using ViT-B/16 and ViT-L/14 models. The evaluation metrics include both text-to-image and image-to-text Recall@K. The best and second-best scores for each method are marked in \textbf{bold} and \underline{underlined}, respectively.}
\vspace{-2mm}
\label{table2}
\end{table*}

We conduct ablation studies to validate the effectiveness of our proposed GOAL framework. Table~\ref{table1} and Table~\ref{table2} present the results on DOCCI and DCI test sets, respectively. We compare four different settings: (1) global fine-tuning with only $\mathcal{L}_{\text{global}}$, (2) local fine-tuning with only $\mathcal{L}_{\text{local}}$, (3) w/o TSL with both $\mathcal{L}_{\text{global}}$ and $\mathcal{L}_{\text{local}}$ without TSL, and (4) our complete GOAL framework with all loss terms.

The results demonstrate the superiority of our framework across all settings. On the DOCCI dataset with ViT-L/14, GOAL achieves 84.37\% R@1 for text-to-image retrieval, surpassing the w/o TSL by 12.87\% (74.75\%), global fine-tuning by 14.01\% (74.00\%), and local fine-tuning by 25.20\% (67.39\%). Similar improvements are observed on the DCI dataset, where GOAL with ViT-L/14 achieves 76.89\% R@1, outperforming the w/o TSL by 15.83\% (66.38\%), global fine-tuning by 16.98\% (65.73\%), and local fine-tuning by 42.70\% (53.88\%). When combined with our proposed TSL method in the complete GOAL framework, we observe consistent improvements across both datasets, demonstrating the effectiveness of our approach.

\renewcommand{\arraystretch}{1.1}
\begin{table}[t]
\begin{center}
\resizebox{\columnwidth}{!}{  
\Large
\begin{tabular}{c|c|ccc|cc}
\toprule
\multirow{2}{*}{Backbone} & \multirow{2}{*}{Method} & \multicolumn{3}{c|}{Loss} & \multicolumn{2}{c}{mAP} \\
\noalign{\vskip 1pt}
\cline{3-7}
\noalign{\vskip 2pt}
& & Global & Local & TSL & T2I & I2T \\
\midrule
\multirow{4}{*}{ViT-B/16} 
& Global fine-tuning & \textcolor{black}{\checkmark} & & & 59.03 & 58.40 \\
& Local fine-tuning & & \textcolor{black}{\checkmark} & & 57.62 & 57.16 \\
& w/o TSL & \textcolor{black}{\checkmark} & \textcolor{black}{\checkmark} & & \underline{60.74} & \underline{59.99} \\
& GOAL & \textcolor{black}{\checkmark} & \textcolor{black}{\checkmark} & \textcolor{black}{\checkmark} & \textbf{63.27} & \textbf{62.63} \\
\midrule
\multirow{4}{*}{ViT-L/14} 
& Global fine-tuning & \textcolor{black}{\checkmark} & & & 65.79 & 64.97 \\
& Local fine-tuning & & \textcolor{black}{\checkmark} & & 62.55 & 62.87 \\
& w/o TSL & \textcolor{black}{\checkmark} & \textcolor{black}{\checkmark} & & \underline{66.55} & \textbf{66.58} \\
& GOAL & \textcolor{black}{\checkmark} & \textcolor{black}{\checkmark} & \textcolor{black}{\checkmark} & \textbf{69.53} & \underline{66.34} \\
\bottomrule
\end{tabular}
}
\end{center}
\vspace{-5mm}
\caption{Comparison of different methods using ViT-B/16 and ViT-L/14 backbones on DOCCI dataset's global and local joint test set. Results show mAP@10 scores for both text-to-image (T2I) and image-to-text (I2T) retrieval tasks. The best and second-best scores for each method are marked in \textbf{bold} and \underline{underlined}, respectively.}
\label{table3}
\end{table}

\renewcommand{\arraystretch}{1.1}
\begin{table}[t]
\begin{center}
\resizebox{\columnwidth}{!}{  
\Large
\begin{tabular}{c|c|ccc|cc}
\toprule
\multirow{2}{*}{Backbone} & \multirow{2}{*}{Method} & \multicolumn{3}{c|}{Loss} & \multicolumn{2}{c}{mAP} \\
\noalign{\vskip 1pt}
\cline{3-7}
\noalign{\vskip 2pt}
& & Global & Local & TSL & T2I & I2T \\
\midrule
\multirow{4}{*}{ViT-B/16} 
& Global fine-tuning & \textcolor{black}{\checkmark} & & & 53.68 & 54.32 \\
& Local fine-tuning & & \textcolor{black}{\checkmark} & & 52.66 & 53.04 \\
& w/o TSL & \textcolor{black}{\checkmark} & \textcolor{black}{\checkmark} & & \underline{56.68} & \underline{56.35} \\
& GOAL & \textcolor{black}{\checkmark} & \textcolor{black}{\checkmark} & \textcolor{black}{\checkmark} & \textbf{57.19} & \textbf{57.35} \\
\midrule
\multirow{4}{*}{ViT-L/14} 
& Global fine-tuning & \textcolor{black}{\checkmark} & & & 55.36 & 58.32 \\
& Local fine-tuning & & \textcolor{black}{\checkmark} & & 52.69 & 54.46 \\
& w/o TSL & \textcolor{black}{\checkmark} & \textcolor{black}{\checkmark} & & \underline{58.60} & \underline{59.85} \\
& GOAL & \textcolor{black}{\checkmark} & \textcolor{black}{\checkmark} & \textcolor{black}{\checkmark} & \textbf{64.77} & \textbf{64.11} \\
\bottomrule
\end{tabular}
}
\end{center}
\vspace{-5mm}
\caption{Comparison of different methods using ViT-B/16 and ViT-L/14 backbones on DCI dataset's global and local joint test set. Results show mAP@10 scores for both text-to-image (T2I) and image-to-text (I2T) retrieval tasks. The best and second-best scores for each method are marked in \textbf{bold} and \underline{underlined}, respectively.}
\label{table4}
\end{table}

We evaluate the methods on a global-local joint test set. Table~\ref{table3} and Table~\ref{table4} present mAP@10 scores for both text-to-image (T2I) and image-to-text (I2T) retrieval tasks on DOCCI and DCI datasets, respectively. The results demonstrate our GOAL framework's capability to effectively handle both global and local feature matching simultaneously. Specifically, on the DOCCI dataset with ViT-L/14, GOAL achieves 69.53\% mAP@10 for T2I, surpassing the w/o TSL (66.55\%) and global fine-tuning (65.79\%) for T2I. Similar improvements are observed on the DCI dataset, where GOAL with ViT-L/14 achieves 64.77\% and 64.11\% for T2I and I2T, respectively, compared to w/o TSL 58.60\% and 59.85\%. These results show that our approach successfully preserves CLIP's global understanding while incorporating local feature matching capabilities, leading to improved performance on both global and local matching tasks.

\subsection{Comparison to the state of the art}
\label{sec:comparison}

\renewcommand{\arraystretch}{1}
\begin{table*}[!t]
\begin{center}
\small  
\begin{tabular}{c|c|cccc|cccc}
\toprule
\multirow{2}{*}{Backbone} & \multirow{2}{*}{Method} & \multicolumn{4}{c|}{Text to Image (Recall@K)} & \multicolumn{4}{c}{Image to Text (Recall@K)} \\
\noalign{\vskip 1pt}
\cline{3-10}
\noalign{\vskip 2pt}
& & R@1 & R@5 & R@25 & R@50 & R@1 & R@5 & R@25 & R@50 \\
\midrule
\multirow{3}{*}{ViT-B/16} 
& Long-CLIP & 61.33 & 80.79 & 91.65 & 94.35 & 60.03 & 81.44 & 92.80 & 95.05 \\
& GOAL DOCCI fine-tuning & \textbf{64.13} & \textbf{82.69} & \textbf{92.95} & \textbf{95.40} & \textbf{65.88} & \textbf{83.44} & \textbf{92.95} & \textbf{95.65} \\
\cmidrule{2-10}
& GOAL DCI fine-tuning & 72.64 & 89.89 & 95.95 & 97.25 & 72.84 & 90.50 & 96.60 & 97.90 \\
\midrule
\multirow{3}{*}{ViT-L/14} 
& Long-CLIP & 67.88 & 83.29 & 91.80 & 94.80 & 64.08 & 84.84 & 93.35 & 95.75 \\
& GOAL DOCCI fine-tuning & \textbf{68.93} & \textbf{85.74} & \textbf{93.95} & \textbf{96.00} & \textbf{68.43} & \textbf{85.99} & \textbf{93.90} & \textbf{96.25} \\
\cmidrule{2-10}
& GOAL DCI fine-tuning & 76.89 & 91.05 & 96.55 & 97.75 & 76.59 & 91.20 & 96.55 & 98.25 \\
\bottomrule
\end{tabular}

\end{center}
\vspace{-5mm}
\caption{Comparison of different methods using ViT-B/16 and ViT-L/14 backbones on DCI dataset.  Results show Text-to-Image and Image-to-Text Recall@K scores in zero-shot setting. The best scores for each method are marked in \textbf{bold}. }
\label{table5}
\end{table*}

\begin{table*}[!t]
\begin{center}
\small
\begin{tabular}{c|c|cccc|cccc}
\toprule
\multirow{2}{*}{Backbone} & \multirow{2}{*}{Method} & \multicolumn{4}{c|}{Text to Image (Recall@K)} & \multicolumn{4}{c}{Image to Text (Recall@K)} \\
\noalign{\vskip 1pt}
\cline{3-10}
\noalign{\vskip 2pt}
& & R@1 & R@5 & R@25 & R@50 & R@1 & R@5 & R@25 & R@50 \\
\midrule
\multirow{3}{*}{ViT-B/16} 
& Long-CLIP & \textbf{71.63} & 92.16 & \textbf{98.90} & \textbf{99.73} & 63.29 & 88.80 & 98.39 & 99.45 \\
& GOAL DCI fine-tuning & 71.22 & \textbf{92.39} & \textbf{98.90} & 99.61 & \textbf{72.18} & \textbf{92.88} & \textbf{98.88} & \textbf{99.55} \\
\cmidrule{2-10}
& GOAL DOCCI fine-tuning & 79.47 & 96.65 & 99.69 & 99.92 & 79.43 & 96.14 & 99.61 & 99.90 \\
\midrule
\multirow{3}{*}{ViT-L/14} 
& Long-CLIP & 78.84 & 95.25 & 99.19 & 99.59 & 66.82 & 91.90 & 99.04 & 99.82 \\
& GOAL DCI fine-tuning & \textbf{79.04} & \textbf{95.78} & \textbf{99.55} & \textbf{99.84} & \textbf{79.16} & \textbf{95.96} & \textbf{99.61} & \textbf{99.90} \\
\cmidrule{2-10}
& GOAL DOCCI fine-tuning & 84.37 & 97.55 & 99.76 & 99.98 & 82.57 & 97.37 & 99.82 & 99.98 \\
\bottomrule
\end{tabular}
\end{center}
\vspace{-5mm}
\caption{Comparison of different methods using ViT-B/16 and ViT-L/14 backbones on DOCCI dataset. Results show Text-to-Image and Image-to-Text Recall@K scores in zero-shot setting. The best scores for each method are marked in \textbf{bold}.}
\label{table6}
\end{table*}

\begin{table*}[!t]
\begin{center}
\small
\begin{tabular}{c|c|cccc}
\toprule
\multirow{2}{*}{Backbone} & \multirow{2}{*}{Method} & \multicolumn{4}{c}{Image to Text (Recall@K)} \\
\noalign{\vskip 1pt}
\cline{3-6}
\noalign{\vskip 2pt}
& & R@1 & R@5 & R@25 & R@50 \\
\midrule
\multirow{4}{*}{ViT-B/16} 
& CLIP & 68.90 & 88.80 & 97.90 & {99.50} \\
& Long-CLIP & 79.20 & 94.80 & 99.10 & \textbf{99.70} \\
& GOAL DOCCI fine-tuning & \underline{81.90} & \underline{95.80} & \textbf{99.40} & \textbf{99.70} \\
& GOAL DCI fine-tuning & \textbf{82.90} & \textbf{96.80} & \textbf{99.40} & \textbf{99.70} \\
\midrule
\multirow{4}{*}{ViT-L/14} 
& CLIP & 68.20 & 88.40 & 97.00 & 98.70 \\
& Long-CLIP & 82.60 & \underline{96.70} & \textbf{99.60} & \textbf{100.00} \\
& GOAL DOCCI fine-tuning & \underline{86.30} & 96.50 & 99.40 & \textbf{100.00} \\
& GOAL DCI fine-tuning & \textbf{89.80} & \textbf{97.80} & \textbf{99.60} & \textbf{100.00} \\
\bottomrule
\end{tabular}

\end{center}
\vspace{-5mm}
\caption{Comparison of different methods using ViT-B/16 and ViT-L/14 backbones on Urban1k dataset. Results show Text-to-Image and Image-to-Text Recall@K scores in zero-shot setting.  The best scores for each method are marked in \textbf{bold}.}
\label{table7}
\end{table*}

We compare our method with Long-CLIP in zero-shot settings across both datasets. For fair comparison, we evaluate fine-tuning methods trained on one dataset and tested on the other (zero-shot), alongside models fine-tuned on the test dataset. In Table~\ref{table5}, our GOAL method fine-tuned on DOCCI outperforms Long-CLIP when tested on the DCI dataset in most metrics, achieving 68.93\% vs 67.88\% in text-to-image R@1 and 68.43\% vs 64.08\% in image-to-text R@1 with ViT-L/14 backbone. The improvement is more pronounced in the ViT-B/16 setting, where our method achieves 64.13\% vs 61.33\% in text-to-image R@1 and 65.88\% vs 60.03\% in image-to-text R@1.

In Table~\ref{table6}, our fine-tuning method on DCI demonstrates strong zero-shot performance compared to Long-CLIP when tested on the DOCCI dataset. With ViT-L/14, GOAL notably outperforms Long-CLIP in higher rank metrics, achieving 95.78\% vs 95.25\% in R@5, 99.55\% vs 99.19\% in R@25 for text-to-image retrieval. The improvement is particularly significant in image-to-text retrieval, where GOAL substantially surpasses Long-CLIP across all metrics, achieving 79.16\% vs 66.82\% in R@1 and 95.96\% vs 91.90\% in R@5. These results demonstrate that our GOAL fine-tuning method exhibits robust generalization capability and superior performance in zero-shot settings across different datasets, with particularly strong improvements in image-to-text retrieval.

\begin{figure*}[!htp]
\centering
\includegraphics[width=\textwidth]{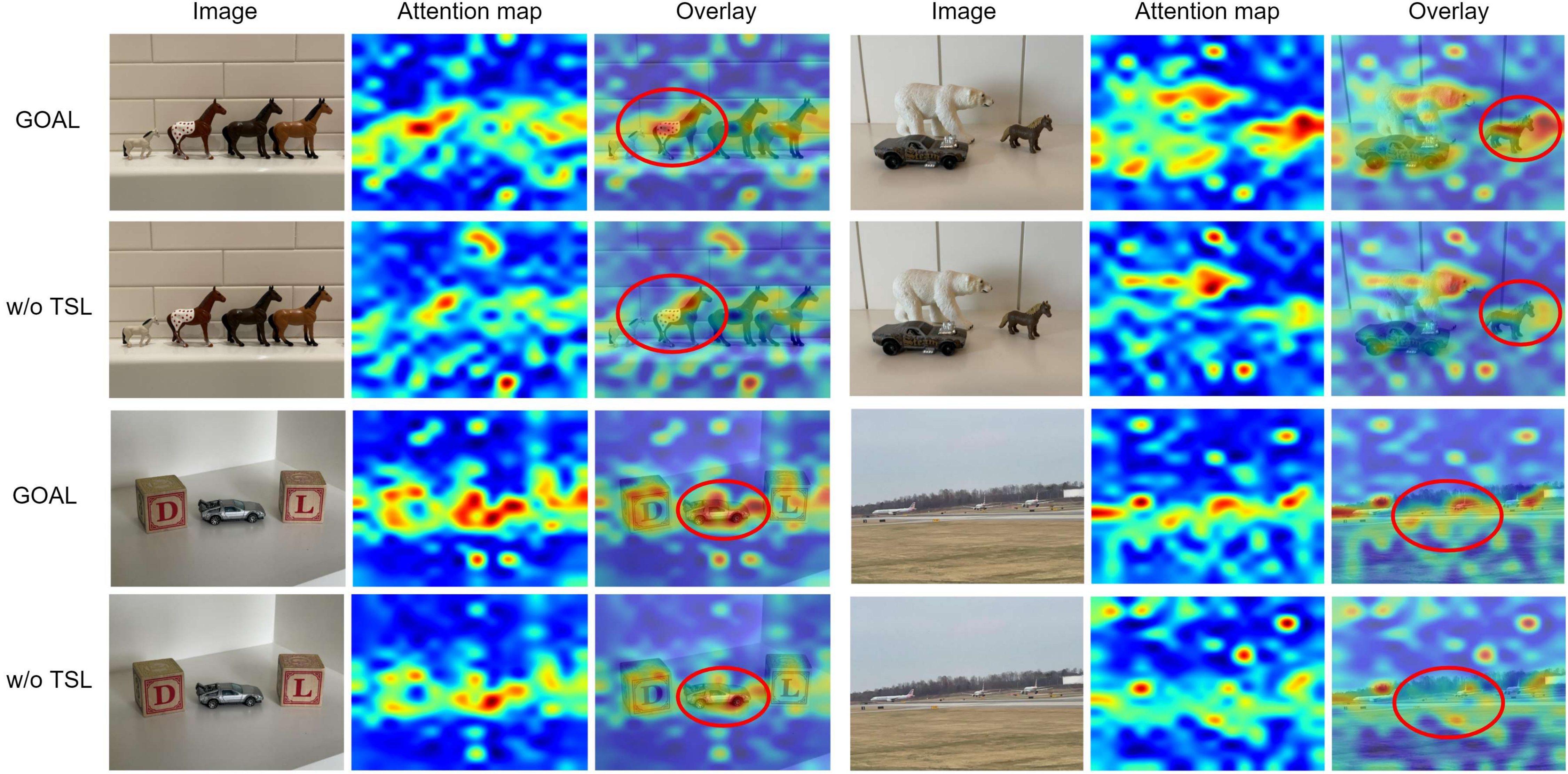}
\caption{Comparison of attention maps generated by GOAL and w/o TSL methods. For each row pair, we present three components: (1) original input image (left), (2) attention heatmap visualization (middle), and (3) overlay of attention on the original image (right). The examples demonstrate how GOAL achieves more focused attention compared to the baseline w/o TSL method. Red circles in the overlay highlight regions where GOAL shows particularly effective attention localization. }
\label{attention_map}
\end{figure*}

Our experiments  on the Urban1k dataset Table~\ref{table7} demonstrate the effectiveness of our approach across fine-tuning methods and pre-trained CLIP. The results show that with the ViT-B/16 backbone, GOAL achieves notable improvements, with GOAL DCI fine-tuning reaching 82.90\% in R@1, surpassing Long-CLIP (79.20\%) and baseline CLIP (68.90\%) by significant margin. The performance gains are even more pronounced with the ViT-L/14 backbone, where GOAL DCI fine-tuning achieves 89.80\% in R@1, outperforming Long-CLIP (82.60\%) and CLIP (68.20\%). Both GOAL variants (DOCCI and DCI fine-tuning) demonstrate competitive performance compared to other fine-tuning methods across recall metrics (R@1, R@5, R@25, R@50), with notable improvements particularly in R@1, which is a crucial metric for retrieval performance. This consistent performance enhancement demonstrates the robustness of our approach in handling image-to-text retrieval tasks, regardless of the backbone architecture used. \\
\indent Additionally, in the supplementary material Sec. \textcolor{red}{B} and Sec. \textcolor{red}{C}, we provide further analysis of our method's ability to preserve global understanding through zero-shot classification experiments on standard benchmarks. We also include extended evaluations comparing our method with BLIP2~\cite{blip2}, and present zero-shot performance results on diverse datasets including COCO~\cite{COCO}, Flickr30k~\cite{Flickr30k}, and ShareGPT4V~\cite{sharegpt4v} to further demonstrate the generalization capabilities of our approach.

\subsection{Qualitative results}
\label{sec:Qualitative results}

We provide qualitative comparisons of attention maps generated by our GOAL and the w/o TSL approach in Fig.~\ref{attention_map}. The visualization~\cite{cam}\cite{gradcam} shows that our GOAL framework captures local details more precisely compared to the w/o TSL. The attention maps clearly show that GOAL consistently focuses on specific objects within the images with higher precision. For instance, in the image containing multiple toy animals, GOAL's attention map shows clear activation across each individual animal figure, while the w/o TSL's attention is more dispersed and partially activated on irrelevant background regions. This enhanced attention behavior demonstrates that GOAL successfully maintains CLIP's global understanding, while incorporating local feature learning through our TSL method. These qualitative results further support our quantitative findings, showing that our fine-tuning method effectively preserves global comprehension while significantly improving the model's ability to attention on local element within the scene.

\section{Conclusion}

In this paper, we have proposed a novel fine-tuning method GOAL that improves CLIP's understanding in image and lengthy text pair datasets. First, Local Image Sentence Matching (LISM) has produced pseudo local pairs through global pairs. Second, Token Similarity based Learning (TSL) has effectively overcome CLIP's limitation of focusing primarily on high-level representations by leveraging attention mechanisms between global and local tokens. Through this research, we have established a foundation for various multi-modal models that perform image-text alignment to effectively learn from lengthy and detailed textual descriptions of images.

\section{Acknowledgment}

This work was supported by the National Research Foundation of Korea(NRF) grant funded by the Korea government(MSIT) (RS-2024-00355008) and the MSIT(Ministry of Science and ICT), Korea, under the Graduate School of Metaverse Convergence support program (IITP-2024-RS-2024-00418847) supervised by the IITP(Institute for Information \& Communications Technology Planning \& Evaluation.
{
    \small
    \bibliographystyle{ieeenat_fullname}

\begin{thebibliography}{41}
\providecommand{\natexlab}[1]{#1}
\providecommand{\url}[1]{\texttt{#1}}
\expandafter\ifx\csname urlstyle\endcsname\relax
  \providecommand{\doi}[1]{doi: #1}\else
  \providecommand{\doi}{doi: \begingroup \urlstyle{rm}\Url}\fi

\bibitem[Chen et~al.(2024)Chen, Lai, Zhang, Wang, Eichner, You, Cao, Zhang, Yang, and Gan]{CLOC}
Hong-You Chen, Zhengfeng Lai, Haotian Zhang, Xinze Wang, Marcin Eichner, Keen You, Meng Cao, Bowen Zhang, Yinfei Yang, and Zhe Gan.
\newblock Contrastive localized language-image pre-training.
\newblock \emph{arXiv preprint arXiv:2410.02746}, 2024.

\bibitem[Chen et~al.(2023)Chen, Li, Dong, Zhang, He, Wang, Zhao, and Lin]{sharegpt4v}
Lin Chen, Jinsong Li, Xiaoyi Dong, Pan Zhang, Conghui He, Jiaqi Wang, Feng Zhao, and Dahua Lin.
\newblock Sharegpt4v: Improving large multi-modal models with better captions.
\newblock \emph{arXiv preprint arXiv:2311.12793}, 2023.

\bibitem[Cui et~al.(2024)Cui, Zhou, Wang, Zhu, and Peng]{reid2}
Zhenyu Cui, Jiahuan Zhou, Xun Wang, Manyu Zhu, and Yuxin Peng.
\newblock Learning continual compatible representation for re-indexing free lifelong person re-identification.
\newblock In \emph{Proceedings of the IEEE/CVF Conference on Computer Vision and Pattern Recognition}, pages 16614--16623, 2024.

\bibitem[Dong et~al.(2023)Dong, Bao, Zheng, Zhang, Chen, Yang, Zeng, Zhang, Yuan, Chen, et~al.]{maskclip}
Xiaoyi Dong, Jianmin Bao, Yinglin Zheng, Ting Zhang, Dongdong Chen, Hao Yang, Ming Zeng, Weiming Zhang, Lu Yuan, Dong Chen, et~al.
\newblock Maskclip: Masked self-distillation advances contrastive language-image pretraining.
\newblock In \emph{Proceedings of the IEEE/CVF Conference on Computer Vision and Pattern Recognition}, pages 10995--11005, 2023.

\bibitem[Dosovitskiy(2020)]{ViT}
Alexey Dosovitskiy.
\newblock An image is worth 16x16 words: Transformers for image recognition at scale.
\newblock \emph{arXiv preprint arXiv:2010.11929}, 2020.

\bibitem[Fu et~al.(2024{\natexlab{a}})Fu, Dai, Luo, Li, Ren, Zhang, Wang, Zhou, Shen, Zhang, et~al.]{Video-mme}
Chaoyou Fu, Yuhan Dai, Yongdong Luo, Lei Li, Shuhuai Ren, Renrui Zhang, Zihan Wang, Chenyu Zhou, Yunhang Shen, Mengdan Zhang, et~al.
\newblock Video-mme: The first-ever comprehensive evaluation benchmark of multi-modal llms in video analysis.
\newblock \emph{arXiv preprint arXiv:2405.21075}, 2024{\natexlab{a}}.

\bibitem[Fu et~al.(2024{\natexlab{b}})Fu, Lin, Long, Shen, Zhao, Zhang, Wang, Yin, Ma, Zheng, et~al.]{vita}
Chaoyou Fu, Haojia Lin, Zuwei Long, Yunhang Shen, Meng Zhao, Yifan Zhang, Xiong Wang, Di Yin, Long Ma, Xiawu Zheng, et~al.
\newblock Vita: Towards open-source interactive omni multimodal llm.
\newblock \emph{arXiv preprint arXiv:2408.05211}, 2024{\natexlab{b}}.

\bibitem[He et~al.(2016)He, Zhang, Ren, and Sun]{resnet}
Kaiming He, Xiangyu Zhang, Shaoqing Ren, and Jian Sun.
\newblock Deep residual learning for image recognition.
\newblock In \emph{Proceedings of the IEEE conference on computer vision and pattern recognition}, pages 770--778, 2016.

\bibitem[Hendrycks et~al.(2021)Hendrycks, Zhao, Basart, Steinhardt, and Song]{imagenet-o}
Dan Hendrycks, Kevin Zhao, Steven Basart, Jacob Steinhardt, and Dawn Song.
\newblock Natural adversarial examples.
\newblock In \emph{Proceedings of the IEEE/CVF conference on computer vision and pattern recognition}, pages 15262--15271, 2021.

\bibitem[Jia et~al.(2021)Jia, Yang, Xia, Chen, Parekh, Pham, Le, Sung, Li, and Duerig]{ALIGN}
Chao Jia, Yinfei Yang, Ye Xia, Yi-Ting Chen, Zarana Parekh, Hieu Pham, Quoc Le, Yun-Hsuan Sung, Zhen Li, and Tom Duerig.
\newblock Scaling up visual and vision-language representation learning with noisy text supervision.
\newblock In \emph{International conference on machine learning}, pages 4904--4916. PMLR, 2021.

\bibitem[Kirillov et~al.(2023)Kirillov, Mintun, Ravi, Mao, Rolland, Gustafson, Xiao, Whitehead, Berg, Lo, et~al.]{SAM}
Alexander Kirillov, Eric Mintun, Nikhila Ravi, Hanzi Mao, Chloe Rolland, Laura Gustafson, Tete Xiao, Spencer Whitehead, Alexander~C Berg, Wan-Yen Lo, et~al.
\newblock Segment anything.
\newblock In \emph{Proceedings of the IEEE/CVF International Conference on Computer Vision}, pages 4015--4026, 2023.

\bibitem[Kottur et~al.(2016)Kottur, Vedantam, Moura, and Parikh]{vis-w2v}
Satwik Kottur, Ramakrishna Vedantam, Jos{\'e}~MF Moura, and Devi Parikh.
\newblock Visual word2vec (vis-w2v): Learning visually grounded word embeddings using abstract scenes.
\newblock In \emph{Proceedings of the IEEE Conference on Computer Vision and Pattern Recognition}, pages 4985--4994, 2016.

\bibitem[Krizhevsky et~al.(2009)Krizhevsky, Hinton, et~al.]{cifar}
Alex Krizhevsky, Geoffrey Hinton, et~al.
\newblock Learning multiple layers of features from tiny images.
\newblock 2009.

\bibitem[Lavoie et~al.(2024)Lavoie, Kirichenko, Ibrahim, Assran, Wilson, Courville, and Ballas]{llip}
Samuel Lavoie, Polina Kirichenko, Mark Ibrahim, Mahmoud Assran, Andrew~Gordon Wilson, Aaron Courville, and Nicolas Ballas.
\newblock Modeling caption diversity in contrastive vision-language pretraining.
\newblock \emph{arXiv preprint arXiv:2405.00740}, 2024.

\bibitem[Li et~al.(2023)Li, Li, Savarese, and Hoi]{blip2}
Junnan Li, Dongxu Li, Silvio Savarese, and Steven Hoi.
\newblock Blip-2: Bootstrapping language-image pre-training with frozen image encoders and large language models.
\newblock In \emph{International conference on machine learning}, pages 19730--19742. PMLR, 2023.

\bibitem[Lin et~al.(2014)Lin, Maire, Belongie, Hays, Perona, Ramanan, Doll{\'a}r, and Zitnick]{COCO}
Tsung-Yi Lin, Michael Maire, Serge Belongie, James Hays, Pietro Perona, Deva Ramanan, Piotr Doll{\'a}r, and C~Lawrence Zitnick.
\newblock Microsoft coco: Common objects in context.
\newblock In \emph{Computer Vision--ECCV 2014: 13th European Conference, Zurich, Switzerland, September 6-12, 2014, Proceedings, Part V 13}, pages 740--755. Springer, 2014.

\bibitem[Liu et~al.(2023)Liu, Li, Wu, and Lee]{Llava}
Haotian Liu, Chunyuan Li, Qingyang Wu, and Yong~Jae Lee.
\newblock Visual instruction tuning.
\newblock \emph{Advances in neural information processing systems}, 36:\penalty0 34892--34916, 2023.

\bibitem[Minderer et~al.(2024)Minderer, Gritsenko, and Houlsby]{OWLv2}
Matthias Minderer, Alexey Gritsenko, and Neil Houlsby.
\newblock Scaling open-vocabulary object detection.
\newblock \emph{Advances in Neural Information Processing Systems}, 36, 2024.

\bibitem[Mo et~al.(2023)Mo, Kim, Lee, and Shin]{sclip}
Sangwoo Mo, Minkyu Kim, Kyungmin Lee, and Jinwoo Shin.
\newblock S-clip: Semi-supervised vision-language learning using few specialist captions.
\newblock \emph{Advances in Neural Information Processing Systems}, 36:\penalty0 61187--61212, 2023.

\bibitem[Onoe et~al.(2024)Onoe, Rane, Berger, Bitton, Cho, Garg, Ku, Parekh, Pont-Tuset, Tanzer, et~al.]{DOCCI}
Yasumasa Onoe, Sunayana Rane, Zachary Berger, Yonatan Bitton, Jaemin Cho, Roopal Garg, Alexander Ku, Zarana Parekh, Jordi Pont-Tuset, Garrett Tanzer, et~al.
\newblock Docci: Descriptions of connected and contrasting images.
\newblock \emph{arXiv preprint arXiv:2404.19753}, 2024.

\bibitem[Radford et~al.(2021)Radford, Kim, Hallacy, Ramesh, Goh, Agarwal, Sastry, Askell, Mishkin, Clark, et~al.]{CLIP}
Alec Radford, Jong~Wook Kim, Chris Hallacy, Aditya Ramesh, Gabriel Goh, Sandhini Agarwal, Girish Sastry, Amanda Askell, Pamela Mishkin, Jack Clark, et~al.
\newblock Learning transferable visual models from natural language supervision.
\newblock In \emph{International conference on machine learning}, pages 8748--8763. PMLR, 2021.

\bibitem[Ren et~al.(2021)Ren, Lin, Zhao, Men, Yang, Zhou, Sun, and Yang]{ren}
Shuhuai Ren, Junyang Lin, Guangxiang Zhao, Rui Men, An Yang, Jingren Zhou, Xu Sun, and Hongxia Yang.
\newblock Learning relation alignment for calibrated cross-modal retrieval.
\newblock \emph{arXiv preprint arXiv:2105.13868}, 2021.

\bibitem[Selvaraju et~al.(2017)Selvaraju, Cogswell, Das, Vedantam, Parikh, and Batra]{gradcam}
Ramprasaath~R Selvaraju, Michael Cogswell, Abhishek Das, Ramakrishna Vedantam, Devi Parikh, and Dhruv Batra.
\newblock Grad-cam: Visual explanations from deep networks via gradient-based localization.
\newblock In \emph{Proceedings of the IEEE international conference on computer vision}, pages 618--626, 2017.

\bibitem[Simonyan and Zisserman(2014)]{vgg}
Karen Simonyan and Andrew Zisserman.
\newblock Very deep convolutional networks for large-scale image recognition.
\newblock \emph{arXiv preprint arXiv:1409.1556}, 2014.

\bibitem[Szegedy et~al.(2016)Szegedy, Vanhoucke, Ioffe, Shlens, and Wojna]{inception}
Christian Szegedy, Vincent Vanhoucke, Sergey Ioffe, Jon Shlens, and Zbigniew Wojna.
\newblock Rethinking the inception architecture for computer vision.
\newblock In \emph{Proceedings of the IEEE conference on computer vision and pattern recognition}, pages 2818--2826, 2016.

\bibitem[Tan et~al.(2024)Tan, Ding, Jiang, Wang, Zhan, and Tao]{reid1}
Wentan Tan, Changxing Ding, Jiayu Jiang, Fei Wang, Yibing Zhan, and Dapeng Tao.
\newblock Harnessing the power of mllms for transferable text-to-image person reid.
\newblock In \emph{Proceedings of the IEEE/CVF Conference on Computer Vision and Pattern Recognition}, pages 17127--17137, 2024.

\bibitem[Urbanek et~al.(2024)Urbanek, Bordes, Astolfi, Williamson, Sharma, and Romero-Soriano]{DCI}
Jack Urbanek, Florian Bordes, Pietro Astolfi, Mary Williamson, Vasu Sharma, and Adriana Romero-Soriano.
\newblock A picture is worth more than 77 text tokens: Evaluating clip-style models on dense captions.
\newblock In \emph{Proceedings of the IEEE/CVF Conference on Computer Vision and Pattern Recognition}, pages 26700--26709, 2024.

\bibitem[Vaswani(2017)]{attention}
A Vaswani.
\newblock Attention is all you need.
\newblock \emph{Advances in Neural Information Processing Systems}, 2017.

\bibitem[Vo et~al.(2019)Vo, Jiang, Sun, Murphy, Li, Fei-Fei, and Hays]{odyssey}
Nam Vo, Lu Jiang, Chen Sun, Kevin Murphy, Li-Jia Li, Li Fei-Fei, and James Hays.
\newblock Composing text and image for image retrieval-an empirical odyssey.
\newblock In \emph{Proceedings of the IEEE/CVF conference on computer vision and pattern recognition}, pages 6439--6448, 2019.

\bibitem[Wang et~al.(2024)Wang, Li, Fu, Xie, Li, Sun, and Ma]{freeze}
Xiong Wang, Yangze Li, Chaoyou Fu, Lei Xie, Ke Li, Xing Sun, and Long Ma.
\newblock Freeze-omni: A smart and low latency speech-to-speech dialogue model with frozen llm.
\newblock \emph{arXiv preprint arXiv:2411.00774}, 2024.

\bibitem[Wang et~al.(2020)Wang, Fang, Wang, and Yang]{ViTAA}
Zhe Wang, Zhiyuan Fang, Jun Wang, and Yezhou Yang.
\newblock Vitaa: Visual-textual attributes alignment in person search by natural language.
\newblock In \emph{Computer Vision--ECCV 2020: 16th European Conference, Glasgow, UK, August 23--28, 2020, Proceedings, Part XII 16}, pages 402--420. Springer, 2020.

\bibitem[Xie et~al.(2017)Xie, Girshick, Doll{\'a}r, Tu, and He]{resnext}
Saining Xie, Ross Girshick, Piotr Doll{\'a}r, Zhuowen Tu, and Kaiming He.
\newblock Aggregated residual transformations for deep neural networks.
\newblock In \emph{Proceedings of the IEEE conference on computer vision and pattern recognition}, pages 1492--1500, 2017.

\bibitem[Yin et~al.(2023)Yin, Fu, Zhao, Li, Sun, Xu, and Chen]{survey}
Shukang Yin, Chaoyou Fu, Sirui Zhao, Ke Li, Xing Sun, Tong Xu, and Enhong Chen.
\newblock A survey on multimodal large language models.
\newblock \emph{arXiv preprint arXiv:2306.13549}, 2023.

\bibitem[Young et~al.(2014)Young, Lai, Hodosh, and Hockenmaier]{Flickr30k}
Peter Young, Alice Lai, Micah Hodosh, and Julia Hockenmaier.
\newblock From image descriptions to visual denotations: New similarity metrics for semantic inference over event descriptions.
\newblock \emph{Transactions of the Association for Computational Linguistics}, 2:\penalty0 67--78, 2014.

\bibitem[Yuan et~al.(2021)Yuan, Chen, Chen, Codella, Dai, Gao, Hu, Huang, Li, Li, et~al.]{Florence}
Lu Yuan, Dongdong Chen, Yi-Ling Chen, Noel Codella, Xiyang Dai, Jianfeng Gao, Houdong Hu, Xuedong Huang, Boxin Li, Chunyuan Li, et~al.
\newblock Florence: A new foundation model for computer vision.
\newblock \emph{arXiv preprint arXiv:2111.11432}, 2021.

\bibitem[Zhai et~al.(2023)Zhai, Mustafa, Kolesnikov, and Beyer]{siglip}
Xiaohua Zhai, Basil Mustafa, Alexander Kolesnikov, and Lucas Beyer.
\newblock Sigmoid loss for language image pre-training.
\newblock In \emph{Proceedings of the IEEE/CVF international conference on computer vision}, pages 11975--11986, 2023.

\bibitem[Zhang et~al.(2025)Zhang, Zhang, Dong, Zang, and Wang]{LongCLIP}
Beichen Zhang, Pan Zhang, Xiaoyi Dong, Yuhang Zang, and Jiaqi Wang.
\newblock Long-clip: Unlocking the long-text capability of clip.
\newblock In \emph{European Conference on Computer Vision}, pages 310--325. Springer, 2025.

\bibitem[Zhang et~al.(2022)Zhang, Zhang, Hu, Chen, Li, Dai, Wang, Yuan, Hwang, and Gao]{GLIPv2}
Haotian Zhang, Pengchuan Zhang, Xiaowei Hu, Yen-Chun Chen, Liunian Li, Xiyang Dai, Lijuan Wang, Lu Yuan, Jenq-Neng Hwang, and Jianfeng Gao.
\newblock Glipv2: Unifying localization and vision-language understanding.
\newblock \emph{Advances in Neural Information Processing Systems}, 35:\penalty0 36067--36080, 2022.

\bibitem[Zheng et~al.(2017)Zheng, Zhang, Sun, Chandraker, Yang, and Tian]{reid3}
Liang Zheng, Hengheng Zhang, Shaoyan Sun, Manmohan Chandraker, Yi Yang, and Qi Tian.
\newblock Person re-identification in the wild.
\newblock In \emph{Proceedings of the IEEE conference on computer vision and pattern recognition}, pages 1367--1376, 2017.

\bibitem[Zhong et~al.(2017)Zhong, Zheng, Cao, and Li]{reid4}
Zhun Zhong, Liang Zheng, Donglin Cao, and Shaozi Li.
\newblock Re-ranking person re-identification with k-reciprocal encoding.
\newblock In \emph{Proceedings of the IEEE conference on computer vision and pattern recognition}, pages 1318--1327, 2017.

\bibitem[Zhou et~al.(2016)Zhou, Khosla, Lapedriza, Oliva, and Torralba]{cam}
Bolei Zhou, Aditya Khosla, Agata Lapedriza, Aude Oliva, and Antonio Torralba.
\newblock Learning deep features for discriminative localization.
\newblock In \emph{Proceedings of the IEEE conference on computer vision and pattern recognition}, pages 2921--2929, 2016.

\end{thebibliography}

}
\clearpage
\setcounter{page}{1}
\maketitlesupplementary

\setcounter{section}{0} 
\renewcommand{\thesection}{\Alph{section}}
\renewcommand{\thesubsection}{\Alph{section}.\arabic{subsection}}

\section{GOAL against Long-CLIP}
\label{GOAL against Long-CLIP}

\begin{table*}[!t]
\centering
\small
\begin{tabular}{c|c|cccc|cccc}
\toprule
\multirow{2}{*}{Backbone} & \multirow{2}{*}{Methods} & \multicolumn{4}{c|}{Text to Image Recall@K} & \multicolumn{4}{c}{Image to Text Recall@K} \\
\noalign{\vskip 1pt}
\cline{3-10}
\noalign{\vskip 2pt}
& & R@1 & R@5 & R@25 & R@50 & R@1 & R@5 & R@25 & R@50 \\
\midrule
\multirow{3}{*}{ViT-B/16} 
& Long-CLIP  & 78.33 & 95.43 & 99.63 & 99.86 & 77.06 & 95.33 & 99.49 & \underline{99.90} \\
& Long-CLIP* & \underline{79.16} & \underline{95.92} & \underline{99.65} & \underline{99.90} & \underline{78.51} & \textbf{96.51} & \textbf{99.67} & \textbf{99.96} \\
& GOAL & \textbf{79.47} & \textbf{96.65} & \textbf{99.69} & \textbf{99.92} & \textbf{79.43} & \underline{96.14} & \underline{99.61} & \underline{99.90} \\
\midrule
\multirow{3}{*}{ViT-L/14}
& Long-CLIP & 83.51 & 97.35 & 99.69 & 99.90 & 81.73 & 96.75 & 99.71 & 99.86 \\
& Long-CLIP* & \textbf{84.80} & \textbf{97.82} & \textbf{99.80} & \underline{99.98} & \textbf{83.45} & \textbf{97.86} & \textbf{99.84} & \underline{99.92} \\
& GOAL & \underline{84.37} & \underline{97.55} & \underline{99.76} & \textbf{99.98} & \underline{82.57} & \underline{97.37} & \underline{99.82} & \textbf{99.98} \\
\bottomrule

\end{tabular}
\captionsetup{width=\linewidth}
\caption{Retrieval performance comparison on DOCCI dataset using different backbones. Long-CLIP* indicates the model fine-tuned with our proposed method, while GOAL represents our complete framework. The best and second-best scores for each method are marked in \textbf{bold} and \underline{underlined}, respectively. }
\vspace{-2mm}
\label{table8}
\end{table*}

We compare our model with Long-CLIP~\cite{LongCLIP} on the DOCCI~\cite{DOCCI} dataset using ViT-B/16 and ViT-L/14~\cite{ViT} backbones in Table~\ref{table8}. The baseline Long-CLIP is first fine-tuned on ShareGPT4V~\cite{sharegpt4v} (1M samples) and then further fine-tuned on DOCCI using standard CLIP~\cite{CLIP} loss. Long-CLIP* follows the same fine-tuned on ShareGPT4V but employs our proposed fine-tuning method on DOCCI, while GOAL is directly fine-tuned on DOCCI from CLIP's pre-trained weights. The results demonstrate a clear performance progression: Long-CLIP* consistently outperforms the baseline Long-CLIP across all metrics, showing the effectiveness of our fine-tuning approach. For example, with ViT-B/16, Long-CLIP* achieves improvements of 1.06\% and 0.51\% in text-to-image retrieval at R@1 and R@5, respectively. Notably, GOAL further surpasses both variants, achieving the best performance across most metrics. With ViT-B/16, GOAL reaches 79.47\% and 79.43\% for R@1 in text-to-image and image-to-text retrieval. This is particularly significant considering that GOAL achieves superior performance while being trained on the DOCCI dataset alone, which is substantially smaller than the combined dataset (ShareGPT4V + DOCCI) used for Long-CLIP. This results demonstrate that our proposed fine-tuning method achieves better performance with significantly reduced data requirements compared to Long-CLIP's fine-tuning approach.

\section{Zero-shot evaluation on short caption datasets}
\label{Zero-shot evaluation on short caption datasets}

\begin{table*}[!t]
\centering
\small
\begin{tabular}{c|c|cccc|cccc}
\toprule
\multirow{2}{*}{Backbone} & \multirow{2}{*}{Methods} & \multicolumn{4}{c|}{Text to Image Recall@K} & \multicolumn{4}{c}{Image to Text Recall@K} \\
\noalign{\vskip 1pt}
\cline{3-10}
\noalign{\vskip 2pt}
& & R@1 & R@5 & R@25 & R@50 & R@1 & R@5 & R@25 & R@50 \\
\midrule
\multirow{4}{*}{ViT-B/16} 
& CLIP & 33.95 & 59.46 & 82.95 & 91.06 & 54.14 & 77.74 & 93.32 & 97.36 \\
& Long-CLIP & \textbf{40.83} & \textbf{66.36} & \textbf{87.42} & \textbf{93.97} & 57.24 & 80.42 & 94.24 & 97.60 \\
& GOAL fine-tuned with DOCCI & 38.86 & 64.36 & 86.22 & 93.28 & \textbf{59.28} & \textbf{81.02} & \underline{94.84} & \underline{97.76} \\
& GOAL fine-tuned with DCI & \underline{39.08} & \underline{65.32} & \underline{86.93} & \underline{93.66} & \underline{57.78} & \underline{80.62} & \textbf{94.90} & \textbf{98.00} \\
\midrule
\multirow{4}{*}{ViT-L/14} 
& CLIP & 37.29 & 61.82 & 84.19 & 91.83 & 57.68 & 80.20 & 94.58 & 97.84 \\
& Long-CLIP & \textbf{46.96} & \textbf{71.89} & \textbf{90.25} & \textbf{95.36} & 63.16 & 84.52 & 96.46 & \textbf{98.66} \\
& GOAL fine-tuned with DOCCI & \underline{46.29} & \underline{70.85} & \underline{89.43} & \underline{95.20} & \textbf{66.50} & \textbf{86.04} & \textbf{96.76} & \underline{98.62} \\
& GOAL fine-tuned with DCI & 45.54 & 70.22 & 89.09 & 94.90 & \underline{64.50} & \underline{85.10} & \underline{96.52} & \underline{98.62} \\
\bottomrule

\end{tabular}
\captionsetup{width=\linewidth}
\caption{Zero-shot evaluation results on COCO test set. Comparison of retrieval performance across different fine-tuning approaches using ViT-B/16 and ViT-L/14 models. The evaluation metrics include both text-to-image and image-to-text Recall@K. The best and second-best scores for each method are marked in \textbf{bold} and \underline{underlined}, respectively.}
\vspace{-2mm}
\label{table9}
\end{table*}

We evaluate our model's zero-shot transfer capabilities on the COCO~\cite{COCO} dataset using both text-to-image and image-to-text retrieval metrics with ViT-B/16 and ViT-L/14 backbones in Table~\ref{table9}. The experimental results demonstrate GOAL's strong performance, particularly when fine-tuned on DOCCI, achieving 66.50\% R@1 in image-to-text retrieval with the ViT-L/14 architecture, surpassing Long-CLIP's 63.16\%. This superior performance extends across higher recall@K values, reaching 86.04\% and 96.76\% for R@5 and R@25 respectively. When fine-tuned on DCI~\cite{DCI}, another detailed caption dataset, GOAL demonstrates consistent performance across all metrics, highlighting its effectiveness across different detailed caption datasets. These comprehensive results validate our model's effectiveness in cross-modal retrieval tasks while maintaining robust adaptability across various datasets.

\begin{table*}[!t]
\centering
\small
\begin{tabular}{c|c|cccc|cccc}
\toprule
\multirow{2}{*}{Backbone} & \multirow{2}{*}{Methods} & \multicolumn{4}{c|}{Text to Image Recall@K} & \multicolumn{4}{c}{Image to Text Recall@K} \\
\noalign{\vskip 1pt}
\cline{3-10}
\noalign{\vskip 2pt}
& & R@1 & R@5 & R@25 & R@50 & R@1 & R@5 & R@25 & R@50 \\
\midrule
\multirow{4}{*}{ViT-B/16} 
& CLIP & 63.20 & 86.30 & 96.48 & 98.52 & 82.90 & \underline{97.20} & 99.40 & \textbf{100.00} \\
& Long-CLIP & \textbf{70.80} & \textbf{90.68} & \textbf{97.74} & \textbf{98.88} & \textbf{85.90} & \textbf{98.50} & \textbf{99.90} & \textbf{100.00} \\
& GOAL fine-tuned with DOCCI & \underline{68.32} & \underline{89.30} & \underline{97.32} & \underline{98.62} & \underline{85.10} & 96.70 & 99.60 & \underline{99.90} \\
& GOAL fine-tuned with DCI & 67.38 & 88.80 & 97.16 & 98.50 & 84.60 & 96.80 & \underline{99.80} & \textbf{100.00} \\
\midrule
\multirow{4}{*}{ViT-L/14} 
& CLIP & 65.38 & 87.36 & 96.84 & 98.30 & 86.40 & 97.50 & \underline{99.90} & \textbf{100.00} \\
& Long-CLIP & \textbf{76.22} & \textbf{93.54} & \underline{98.36} & \underline{99.28} & \underline{90.00} & \textbf{98.90} & \underline{99.90} & \textbf{100.00} \\
& GOAL fine-tuned with DOCCI & \underline{74.76} & \underline{92.66} & \textbf{98.44} & \textbf{99.32} & \textbf{90.80} & \underline{98.80} & \underline{99.90} & \textbf{100.00} \\
& GOAL fine-tuned with DCI & 73.76 & 91.92 & 98.22 & 99.20 & 89.10 & 98.30 & \textbf{100.00} & \textbf{100.00} \\
\bottomrule

\end{tabular}
\captionsetup{width=\linewidth}
\caption{Zero-shot evaluation results on Flickr30K test set. Comparison of retrieval performance across different fine-tuning approaches using ViT-B/16 and ViT-L/14 models. The evaluation metrics include both text-to-image and image-to-text Recall@K. The best and second-best scores for each method are marked in \textbf{bold} and \underline{underlined}, respectively.}
\vspace{-2mm}
\label{table10}
\end{table*}

We further validate our model's zero-shot transfer capabilities on the Flickr30K~\cite{Flickr30k} using both text-to-image and image-to-text retrieval metrics with ViT-B/16 and ViT-L/14 backbones in Table~\ref{table10}. The experimental results demonstrate GOAL's strong performance, particularly when fine-tuned on DOCCI with the ViT-L/14 architecture, achieving 90.80\% R@1 in image-to-text retrieval and maintaining high performance with 98.80\% and 99.90\% for R@5 and R@25 respectively. In text-to-image retrieval, GOAL fine-tuned on DOCCI demonstrates robust performance, achieving 74.76\% R@1 and 92.66\% R@5. Furthermore, when fine-tuned on DCI, another detailed caption dataset, GOAL maintains consistent performance across all metrics, showing comparable results with 73.76\% and 91.92\% for R@1 and R@5 in text-to-image retrieval, and 89.10\% and 98.30\% for R@1 and R@5 in image-to-text retrieval. These comprehensive results demonstrate our model's effectiveness in cross-modal retrieval tasks while maintaining robust performance across different detailed caption datasets.

\section{Further analysis on GOAL}
\label{Further analysis on GOAL}

\begin{table*}[!t]
\centering
\small
\begin{tabular}{c|c|cccc|cccc}
\toprule
\multirow{2}{*}{Backbone} & \multirow{2}{*}{Methods} & \multicolumn{4}{c|}{Text to Image Recall@K} & \multicolumn{4}{c}{Image to Text Recall@K} \\
\noalign{\vskip 1pt}
\cline{3-10}
\noalign{\vskip 2pt}
& & R@1 & R@5 & R@25 & R@50 & R@1 & R@5 & R@25 & R@50 \\
\midrule
\multirow{3}{*}{ViT-B/16} 
& CLIP  & 61.12 & 83.82 & 95.84 & 98.42 & 62.24 & 82.32 & 95.00 & 97.74 \\
& CLIP+LongCLIP & \underline{66.86} & \underline{88.72} & \underline{97.56} & \underline{99.20} & \underline{75.56} & \underline{93.36} & \underline{98.78} & \underline{99.62} \\
& CLIP+GOAL & \textbf{79.50} & \textbf{94.82} & \textbf{99.34} & \textbf{99.74} & \textbf{85.44} & \textbf{97.12} & \textbf{99.62} & \textbf{99.84} \\
\midrule
\multirow{3}{*}{ViT-L/14}
& CLIP & 53.72 & 76.40 & 91.28 & 95.60 & 62.70 & 81.78 & 93.78 & 96.64 \\
& CLIP+LongCLIP & \underline{66.85} & \underline{88.80} & \underline{97.62} & \underline{99.14} & \underline{73.84} & \underline{91.44} & \underline{98.50} & \underline{99.48} \\
& CLIP+GOAL & \textbf{85.48} & \textbf{96.84} & \textbf{99.66} & \textbf{99.86} & \textbf{88.62} & \textbf{97.88} & \textbf{99.76} & \textbf{99.92} \\
\bottomrule

\end{tabular}
\captionsetup{width=\linewidth}
\caption{ Comparison of retrieval performance on a test set of 5,000 randomly sampled images from ShareGPT4V. All models were fine-tuned on the DOCCI dataset. The best and second-best scores for each method are marked in \textbf{bold} and \underline{underlined}, respectively. }
\vspace{-2mm}
\label{table11}
\end{table*}
\begin{table*}[!t]
\centering
\small
\begin{tabular}{c|c|cccc|cccc}
\toprule
\multirow{2}{*}{Backbone} & \multirow{2}{*}{Methods} & \multicolumn{4}{c|}{Text to Image Recall@K} & \multicolumn{4}{c}{Image to Text Recall@K} \\
\noalign{\vskip 1pt}
\cline{3-10}
\noalign{\vskip 2pt}
& & R@1 & R@5 & R@25 & R@50 & R@1 & R@5 & R@25 & R@50 \\
\midrule
\multirow{3}{*}{ViT-B/16} 
& CLIP  & 53.30 & 76.70 & 91.50 & 95.40 & 68.90 & 88.80 & 97.90 & 99.95 \\
& CLIP+LongCLIP & \underline{61.30} & \underline{83.90} & \underline{96.80} & \underline{98.80} & \underline{63.60} & \underline{85.90} & \underline{96.80} & \underline{99.00} \\
& CLIP+GOAL & \textbf{73.20} & \textbf{92.70} & \textbf{98.30} & \textbf{99.40} & \textbf{81.90} & \textbf{95.80} & \textbf{99.40} & \textbf{99.70} \\
\midrule
\multirow{3}{*}{ViT-L/14}
& CLIP & 53.90 & 78.40 & 92.20 & 95.80 & 68.20 & 88.40 & 97.00 & 98.80 \\
& CLIP+LongCLIP & \underline{60.60} & \underline{83.00} & \underline{96.00} & \underline{98.60} & \underline{70.20} & \underline{89.80} & \underline{97.50} & \underline{98.70} \\
& CLIP+GOAL & \textbf{83.00} & \textbf{95.40} & \textbf{99.70} & \textbf{99.90} & \textbf{86.30} & \textbf{96.50} & \textbf{99.40} & \textbf{100.00} \\
\bottomrule

\end{tabular}
\captionsetup{width=\linewidth}
\caption{ Comparison of text-to-image and image-to-text retrieval performance on the Urban1k test set. All models were fine-tuned on DOCCI dataset. The best and second-best scores for each method are marked in \textbf{bold} and \underline{underlined}, respectively. }
\vspace{-2mm}
\label{table12}
\end{table*}
\begin{table}[!t]
\centering
\resizebox{\columnwidth}{!}{
\begin{tabular}{c|c|c|c|c}
\toprule
\multirow{1}{*}{\begin{tabular}[c]{@{}c@{}}Backbone\end{tabular}} & 
\multirow{1}{*}{\begin{tabular}[c]{@{}c@{}}Methods\end{tabular}} & 
CIFAR10 & CIFAR100 & ImageNet-O \\ \midrule
\multirow{2}{*}{ViT-B/16} & CLIP+LongCLIP & 85.52 & 54.94 & 36.00 \\
 & CLIP+GOAL & \textbf{87.54} & \textbf{59.70} & \textbf{40.35} \\ \bottomrule
\end{tabular}
}
\caption{ Zero-shot classification accuracy comparison between
CLIP fine-tuned with Long-CLIP method and CLIP fine-tuned with
GOAL method on CIFAR and ImageNet-O datasets. The best scores for each method are marked in \textbf{bold}.}
\vspace{-2mm}
\label{table13}
\end{table}

We evaluate the effectiveness of our proposed GOAL method against the baseline CLIP model and Long-CLIP fine-tuning approach. While our previous experiments in Sec.~\ref{GOAL against Long-CLIP} demonstrated the benefits of applying our method on top of Long-CLIP fine-tuning, here we present a direct comparison between different fine-tuning strategies applied to the original CLIP model. For Long-CLIP fine-tuning, which requires short captions that are not originally included in DOCCI, we generated concise one-sentence descriptions using LLaVA-1.5-7b~\cite{Llava} to create the necessary short captions. The dataset containing these generated short captions is available in our GitHub\footnote{\label{2}\href{https://github.com/PerceptualAI-Lab/GOAL/tree/main/datasets}{https://github.com/PerceptualAI-Lab/GOAL/tree/main/datasets}}.

 Table~\ref{table11} presents the text-to-image and image-to-text retrieval results on a test set of 5,000 samples randomly selected from ShareGPT4V, with all models fine-tuned on the DOCCI dataset. This randomly sampled test set is also available in our GitHub\textsuperscript{\textcolor{red}{2}}. Our proposed GOAL method demonstrates substantial improvements over the Long-CLIP approach. For text-to-image retrieval, GOAL surpasses Long-CLIP by 18.91\% with ViT-B/16 and by 27.87\% with ViT-L/14 in R@1 scores. For image-to-text retrieval, GOAL outperforms Long-CLIP by 13.07\% with ViT-B/16 and by 20.02\% with ViT-L/14 in R@1 scores. This consistent improvement across all retrieval metrics indicates enhanced performance at various retrieval levels. These results confirm that our GOAL fine-tuning approach more effectively adapts the CLIP model, showing strong improvements across both the ViT-B/16 and ViT-L/14 backbones.

 We further evaluate the performance of our models on the Urban1k test set, as shown in Table~\ref{table12}. Similar to the results observed on the ShareGPT4V test set, GOAL consistently outperforms both the baseline CLIP and Long-CLIP fine-tuning approaches across all metrics. With the ViT-B/16 backbone, CLIP+GOAL achieves 73.20\% and 81.90\% R@1 for text-to-image and image-to-text retrieval, exceeding Long-CLIP by 19.41\% and 28.77\%, respectively. The performance gap widens further with the ViT-L/14 backbone, where GOAL achieves impressive R@1 scores of 83.00\% for text-to-image and 86.30\% for image-to-text retrieval, surpassing Long-CLIP by  36.96\% and 22.93\%. These results on Urban1k~\cite{LongCLIP} further validate that our approach generalizes well across different datasets, demonstrating consistent improvements regardless of the test data distribution.

We also evaluate our proposed GOAL method's ability to preserve global visual understanding capabilities, such as those required for classification tasks. Table~\ref{table13} presents the zero-shot classification performance of models fine-tuned on the DOCCI dataset. When evaluated on CIFAR10~\cite{cifar}, CIFAR100~\cite{cifar}, and ImageNet-O~\cite{imagenet-o} datasets, CLIP fine-tuned with the GOAL method consistently outperforms the Long-CLIP approach. Specifically, GOAL achieves 87.54\% accuracy on CIFAR10, 59.70\% on CIFAR100, and 40.35\% on ImageNet-O, showing improvements of 2.36\%, 8.66\%, and 12.08\%, respectively over Long-CLIP. These results suggest that the GOAL method effectively preserves the model's global understanding capabilities while adapting to new tasks. This demonstrates that GOAL offers a balanced approach that maintains the model's general visual representation abilities even after fine-tuning.

\section{Experiments on different backbone}
\label{Experiments on different backbone}

\begin{table}[!t]
\centering
\resizebox{\columnwidth}{!}{
\begin{tabular}{c|c|cc|cc}
\toprule
\multirow{2}{*}{\begin{tabular}[c]{@{}c@{}}Backbone\end{tabular}} & \multirow{2}{*}{\begin{tabular}[c]{@{}c@{}}Method\end{tabular}} & \multicolumn{2}{c|}{T2I} & \multicolumn{2}{c}{I2T} \\ \cmidrule{3-6}
 &  & R@1 & R@5 & R@1 & R@5 \\ \midrule
\multirow{2}{*}{BLIP2-Giant} & BLIP2+CLIP & 23.45 & 54.96 & 26.16 & 57.53 \\
 & BLIP2+GOAL & \textbf{64.63} & \textbf{90.02} & \textbf{61.86} & \textbf{88.47} \\ \bottomrule
\end{tabular}
}
\caption{Cross-modal retrieval performance comparison on DOCCI dataset between BLIP2 fine-tuned with CLIP method and BLIP2 fine-tuned with GOAL method. The best scores for each method are marked in \textbf{bold}.}
\vspace{-2mm}
\label{table14}
\end{table}
\begin{table}[!t]
\centering
\resizebox{\columnwidth}{!}{
\begin{tabular}{c|c|cc|cc}
\toprule
\multirow{2}{*}{\begin{tabular}[c]{@{}c@{}}Backbone\end{tabular}} & \multirow{2}{*}{\begin{tabular}[c]{@{}c@{}}Method\end{tabular}} & \multicolumn{2}{c|}{T2I} & \multicolumn{2}{c}{I2T} \\ \cmidrule{3-6}
 &  & R@1 & R@5 & R@1 & R@5 \\ \midrule
\multirow{2}{*}{BLIP2-Giant} & BLIP2+CLIP & 22.81 & 52.33 & 20.11 & 50.28 \\
 & BLIP2+GOAL & \textbf{50.88} & \textbf{77.49} & \textbf{50.38} & \textbf{77.49} \\ \bottomrule
\end{tabular}
}
\caption{Cross-modal retrieval performance comparison on DCI dataset between BLIP2 fine-tuned with CLIP method and BLIP2 fine-tuned with GOAL method. The best scores for each method are marked in \textbf{bold}.}
\vspace{-6mm}
\label{table15}
\end{table}

We extend our evaluation to explore GOAL's effectiveness when applied to SOTA vision-language models. Tables~\ref{table14} and~\ref{table15} present the cross-modal retrieval performance comparison between BLIP2~\cite{blip2} fine-tuned with standard CLIP-style and our proposed GOAL method on the DOCCI and DCI datasets, respectively. On the DOCCI dataset, BLIP2+GOAL significantly outperforms BLIP2+CLIP, achieving 64.63\% and 61.86\% R@1 for text-to-image and image-to-text retrieval. Similarly on the DCI dataset, BLIP2+GOAL reaches 50.88\% and 50.38\% R@1. We want to note that our GOAL method is model-agnostic and can be applied to state-of-the-art vision-language models for efficient fine-tuning toward better understanding of images with lengthy text descriptions, as shown in these tables. These significant performance improvements across different model architectures confirm the broad applicability and effectiveness of our proposed method.

\section{Retrieval qualitative results}
\label{Retrieval qualitative results}
\begin{figure*}[!t]
\ContinuedFloat*
\begin{minipage}{\textwidth}
\centering
\includegraphics[width=\textwidth]{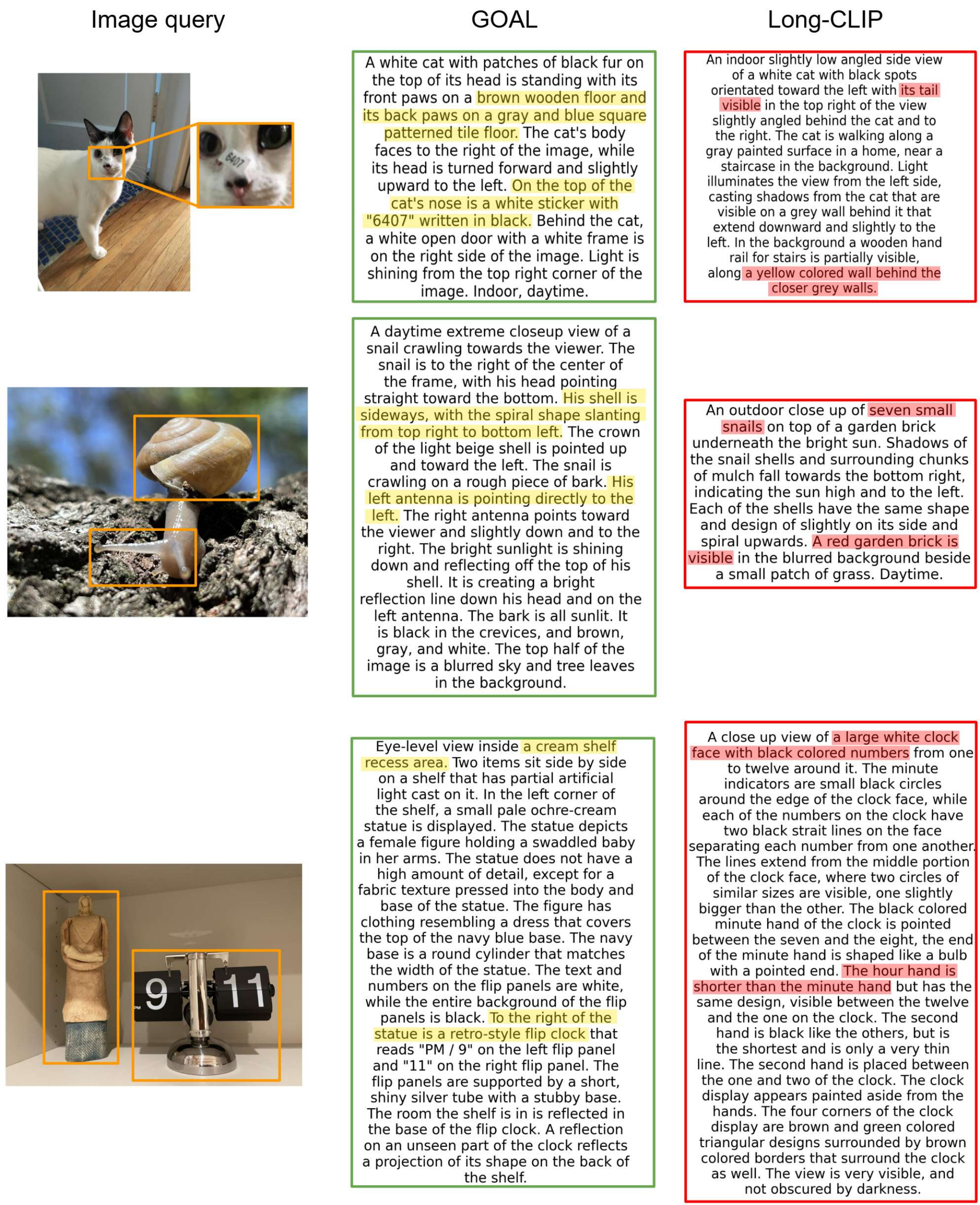}

\label{retrieval results}
\end{minipage}
\end{figure*}
\begin{figure*}[!t]
\ContinuedFloat
\begin{minipage}{\textwidth}
\centering
\includegraphics[width=\textwidth]{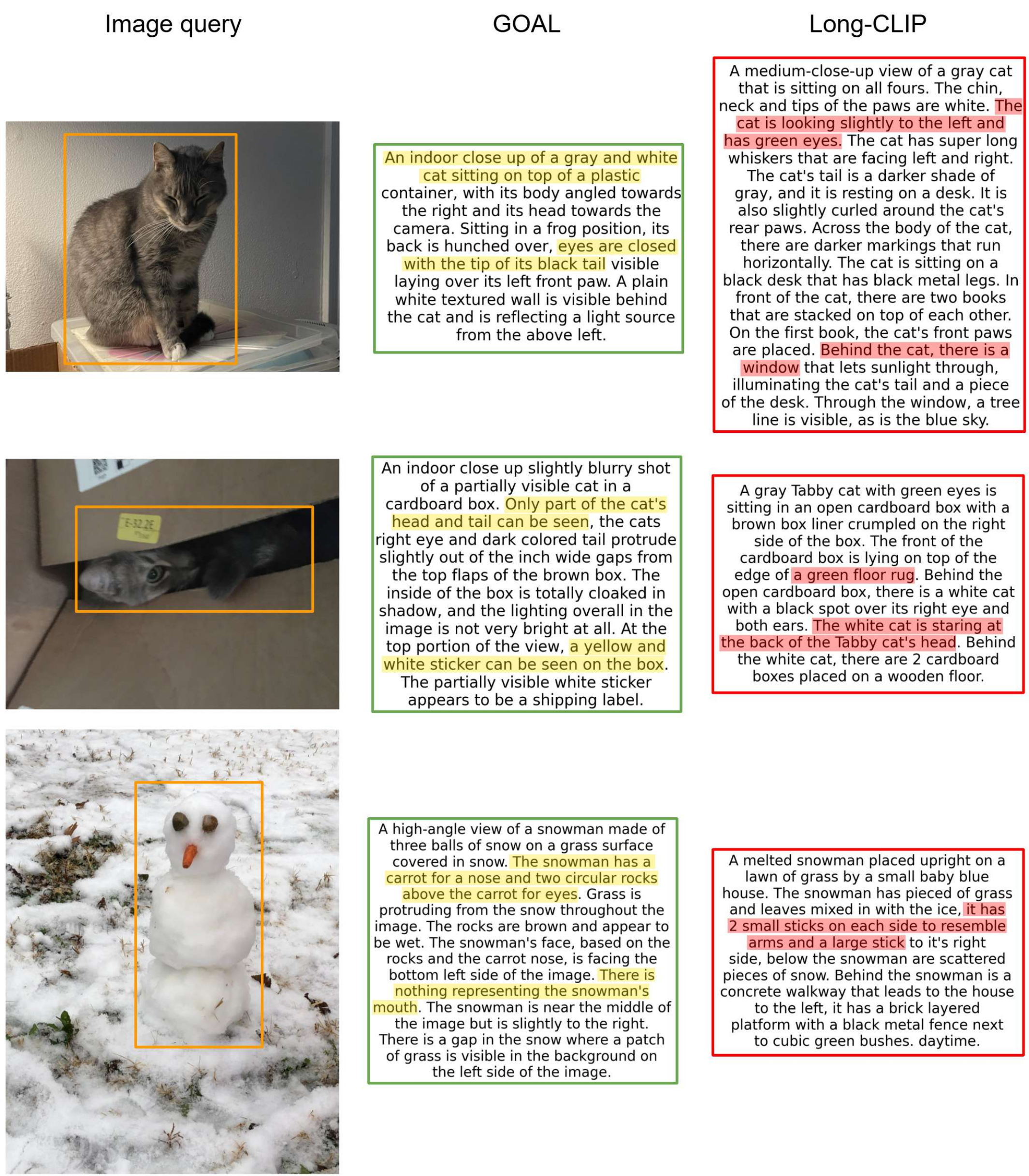}
\caption{Qualitative comparison of image-text retrieval results between GOAL (middle column) and Long-CLIP (right column). The retrieved descriptions demonstrate GOAL's superior ability to capture fine-grained details and diverse scene elements across indoor and outdoor environments, while maintaining semantic coherence in lengthy descriptions. Query images are shown in the left column.}
\end{minipage}
\end{figure*}

We demonstrate the effectiveness of GOAL through qualitative comparison of correctly and incorrectly retrieved captions based on image queries in Fig.~\ref{retrieval results}. The green boxes show correctly retrieved results, while the red boxes show the incorrectly retrieved results. GOAL consistently retrieves more precise and detailed descriptions across various scenarios. In the first row example, GOAL accurately captures specific details like the ``6407" sticker, the distinct floor transitions (wooden and tiled), and precise spatial relationships of architectural elements, which are made possible through TSL's local element attention mechanism. Similarly, in the second row, GOAL correctly matches descriptions containing fine-grained details including antennae orientation and shell positioning, along with precise environmental lighting conditions. In contrast, Long-CLIP (red boxes), trained using the approach described in Sec.~\ref{Further analysis on GOAL}, fails to retrieve accurate descriptions, instead returning more general descriptions that miss crucial visual details and spatial relationships. These results effectively demonstrate that GOAL provides enhanced capability in processing and understanding lengthy and detailed captions, making it a key advantage over Long-CLIP implementations.


\end{document}